\documentclass[aoas]{imsart}

\RequirePackage{amsthm,amsmath,amsfonts,amssymb}
\RequirePackage[authoryear]{natbib}
\RequirePackage[colorlinks,citecolor=blue,urlcolor=blue]{hyperref}%% uncomment this for coloring bibliography citations and linked URLs
\RequirePackage{graphicx}%% uncomment this for including figures
\usepackage{enumitem}
\usepackage[utf8]{inputenc} % allow utf-8 input
\usepackage[T1]{fontenc}    % use 8-bit T1 fonts
\usepackage{hyperref}       % hyperlinks

\usepackage{booktabs}       % professional-quality tables
\usepackage{tabularx}
\usepackage{multirow}
\newcolumntype{Y}{>{\centering\arraybackslash}X}
\newcolumntype{Z}{>{\raggedleft\arraybackslash}X}
\usepackage{amsfonts}       % blackboard math symbols
\usepackage{nicefrac}       % compact symbols for 1/2, etc.
\usepackage{microtype}      % microtypography
\usepackage[dvipsnames]{xcolor}         % colors
\usepackage{amsmath,amssymb,amsthm}
\usepackage{bbold}
\usepackage{colortbl}
\usepackage{mathrsfs,mathtools}
\usepackage{subfigure}
\usepackage{thm-restate}
\usepackage{wrapfig}
\usepackage{caption}
\usepackage[capitalize]{cleveref}
\usepackage{tikz}
\usetikzlibrary{calc, fadings, decorations, shapes, positioning, arrows}
\usepackage{pifont}
\usepackage{tcolorbox}
\usepackage{soul}
\usepackage[parfill]{parskip}
\definecolor{salmon}{RGB}{234,153,153}
\definecolor{cornflowerblue}{RGB}{100,149,237}
 \hypersetup{
     colorlinks,
     linkcolor={magenta},
    filecolor={cyan},
	citecolor={blue},
    urlcolor={purple}
 }
\usepackage{makecell}
\definecolor{darkgreen}{rgb}{0.0, 0.5, 0.0}
\usepackage{tikz}
\startlocaldefs
\theoremstyle{plain}
\newtheorem{theorem}{Theorem}[section]

\theoremstyle{definition}

\endlocaldefs

\begin{document}

\begin{frontmatter}
\title{Robust Reinforcement Learning from Human Feedback for
Large Language Models Fine-Tuning}
%\title{A sample article title with some additional note\thanksref{T1}}
\runtitle{Robust RLHF for LLMs Fine-Tuning}
%\thankstext{T1}{A sample of additional note to the title.}

\begin{aug}
%%%%%%%%%%%%%%%%%%%%%%%%%%%%%%%%%%%%%%%%%%%%%%%
%% Only one address is permitted per author. %%
%% Only division, organization and e-mail is %%
%% included in the address.                  %%
%% Additional information (such as           %%
%% indicating the corresponding author) can  %%
%% be included in the Acknowledgments        %%
%% section if necessary.                     %%
\author[A]{\fnms{Kai}~\snm{Ye} \thanksref{t1}\ead[label=e1]{k.ye1@lse.ac.uk}}
\author[B]{\fnms{Hongyi}~\snm{Zhou} \thanksref{t1}\ead[label=e2]{zhou-hy21@mails.tsinghua.edu.cn}}
\author[C]{\fnms{Jin}~\snm{Zhu} \thanksref{t1}\ead[label=e3]{
j.zhu.7@bham.ac.uk}}
\author[D]{\fnms{Francesco}~\snm{Quinzan}\thanksref{t2}\ead[label=e4]{francesco.quinzan@eng.ox.ac.uk}}
\and
\author[A]{\fnms{Chengchun}~\snm{Shi}\thanksref{t2}\ead[label=e5]{c.shi7@lse.ac.uk}}

%%%%%%%%%%%%%%%%%%%%%%%%%%%%%%%%%%%%%%%%%%%%%%
%% Addresses                                %%
%%%%%%%%%%%%%%%%%%%%%%%%%%%%%%%%%%%%%%%%%%%%%%
\address[A]{Department of Statistics, LSE\printead[presep={,\ }]{e1,e5}}

\address[B]{Department of Mathematics, Tsinghua University\printead[presep={,\ }]{e2}}
\address[C]{School of Mathematics,
University of Birmingham\printead[presep={,\ }]{e3}}

\address[D]{3Department of Engineering Science, University of Oxford\printead[presep={,\ }]{e4}}
\thankstext{t1}{These authors contributed equally.}
\thankstext{t2}{Corresponding authors.}
\end{aug}

\begin{abstract}
Reinforcement learning from human feedback (RLHF) has emerged as a critical technique for aligning the output of large language models (LLMs) with human preferences. It typically learns a reward function from human preference data to measure the quality of an LLM's response to a given question. Most existing algorithms model this reward using the Bradley-Terry model, which relies on assumptions about human preferences that may not reflect the complexity and variability of real-world judgments.  In this paper, we propose a robust algorithm to enhance the performance of existing approaches under such model misspecifications. Theoretically, our algorithm reduces the variance of reward and policy estimators, leading to improved regret bounds. Empirical evaluations on LLM benchmark datasets demonstrate that the proposed algorithm consistently outperforms existing methods, with 53-98\% of responses being favored over baselines on the Anthropic Helpful and Harmless dataset. The code is available at \textit{\url{https://github.com/VRPO/VRPO}}.
\end{abstract}


\end{frontmatter}
\section{Introduction}
\begin{figure}[t]
    \centering
    \includegraphics[width=0.95\linewidth]{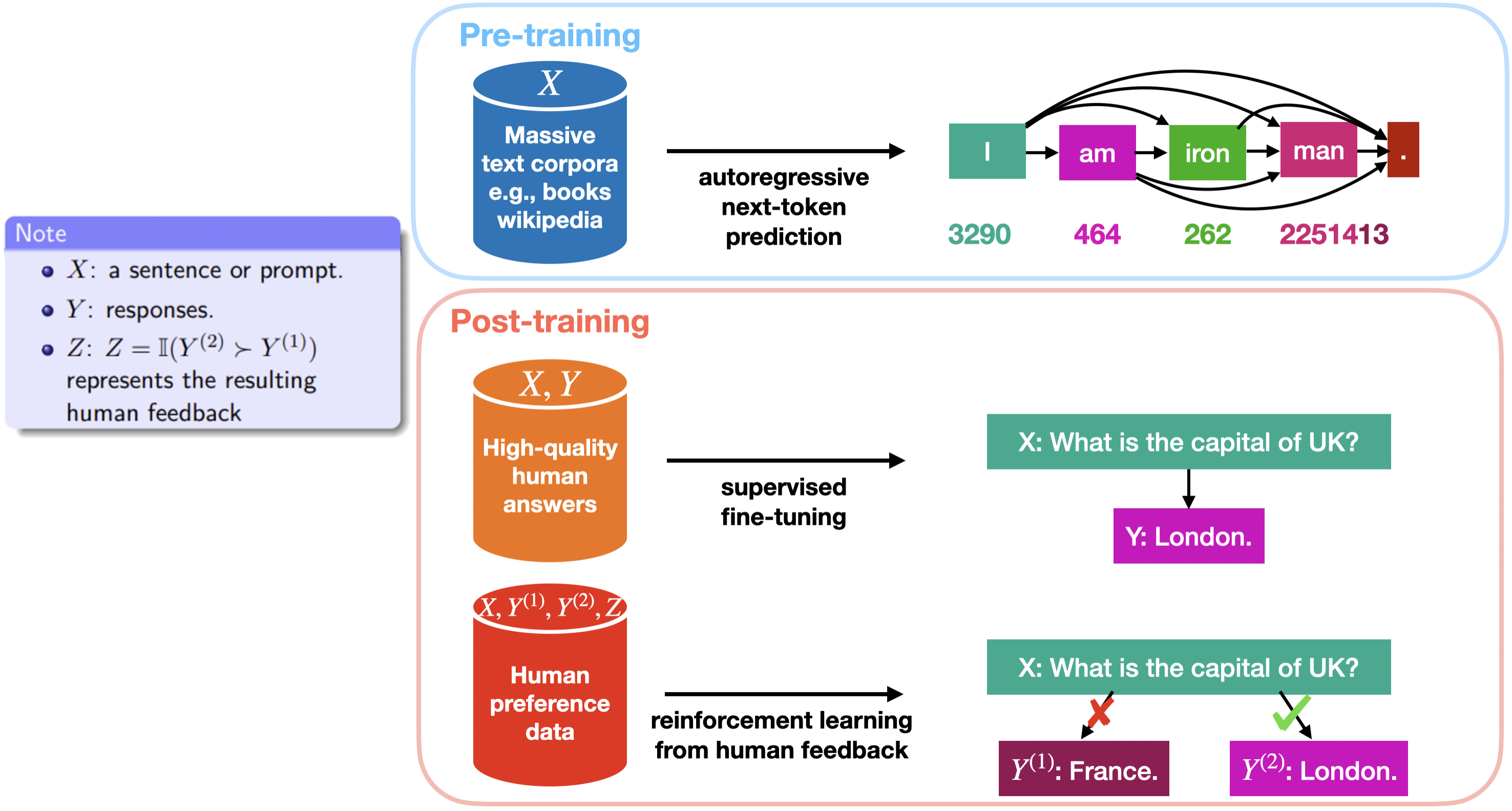}
    \caption{\textbf{Training of LLMs}. (a) The upper panel visualizes autoregressive next-token prediction in pre-training, where each token (e.g., a word or punctuation mark) is encoded into a numerical integer, and its probability depends on all preceding tokens through the transformer architecture. (b) The bottom panel visualizes SFT and RLHF in post-training. Supervised fine-tuning (SFT) fine-tunes the model on a small dataset of high-quality human-written answers to align its outputs with these answers. RLHF employs RL based on human preference, specifying which of the two candidate answers is preferred.}\label{fig:trainLLM}
\end{figure}

We have definitely entered an era of generative artificial intelligence, where large language models (LLMs) are increasingly shaping our daily lives. Their impact is everywhere -- transforming how people access information for learning, solve problems at work, and manage everyday activities. The training of LLMs consists of two stages: (i) a pre-training stage, in which models learn linguistic patterns from large text corpora, and (ii) a post-training stage, which aligns model outputs with human values such as helpfulness, honesty, and harmlessness \citep{askell2021general}. Pre-training is performed using autoregressive next-token prediction, where the model learns the conditional probability distribution of a token (a word, subword, or punctuation) given all preceding tokens, parameterized by the transformer architecture \citep{vaswani2017attention}. Post-training then fine-tunes the model parameters using a smaller number of data samples via supervised fine-tuning \citep[SFT,][]{touvron_2023_llamaopenefficientfoundation} or reinforcement learning from human feedback \citep[RLHF,][]{christiano2017deep}. While pre-training enables LLMs to mimic human language, it may produce factually incorrect responses. Post-training is therefore essential for real-world deployment, as it steers models toward generating responses that are not only human-like but also useful and reliable. A graphical visualization of these two stages is provided in Figure~\ref{fig:trainLLM}.

This paper focuses on RLHF, which employs reinforcement learning \citep[RL,][]{sutton2018reinforcement} for post-training and has recently achieved remarkable success in aligning LLMs with human preferences \citep{bai2022training, ouyang2022training,GPT4TR,shao2024deepseekmath,yang2025qwen3}. Traditional RL algorithms rely on explicitly defined reward functions \citep{agarwal2019reinforcement,levine2020offline,uehara2022review}, but specifying such functions for LLMs is notoriously challenging due to the subtlety and variability of human values \citep{Bertrand_2023_Elo, munos2024nash}. RLHF addresses this limitation by leveraging direct human feedback, such as pairwise comparisons or rankings — which are easier to elicit and more aligned with human intuition. This approach enables LLMs to produce responses that better reflect nuanced human preferences. Existing RLHF algorithms typically require to specify a human preference model. Among those available, the most widely adopted is the Bradley-Terry (BT) model \citep{bradley1952rank}, preferred for its computational tractability and theoretical convenience. 

Models like BT, however, rely on the reward-based preference condition (see Equation \eqref{eqn:peferencereward} in Section \ref{sec:preliminary} for its detailed definition), which entails various unrealistic assumptions on human preferences. One of these assumptions is transitivity -- human preferences are logically ordered ($Y^{(1)}\succ Y^{(2)}\succ Y^{(3)}$ implies $Y^{(1)}\succ Y^{(3)}$\footnote{Here, $Y^{(1)} \succ Y^{(2)}$ indicates that response $Y^{(1)}$ is preferred to response $Y^{(2)}$.}). However, empirical evidence consistently demonstrates that human preferences are inherently intransitive \citep{May_1954_IntransitivityUA, Tversky1969IntransitivityOP, Gardener_1970_Mathematicalgames}. Another unrealistic assumption is context-independence, i.e., preferences between two responses are based solely on the prompts and responses themselves, overlooking the dynamic nature of human-AI interactions \citep{Michaud2020UnderstandingLR, Milano_2021_Ethical, Lindner_2022_HumansAN}. A third assumption is the prefect rationality of the users providing feedback, since humans often provide inconsistent and stochastic preferences \citep{Agranov_2015_StochasticCA}. For these reasons, reward-based preference models like BT may be misspecified in practice.  A possible solution to this problem is to use more general preference models (see Section \ref{sec:relatedworks}). However, more complex models could increase the computational cost and slow the policy learning \citep{zhang2024general}.

Moreover, even when the BT model assumption holds, the reward model itself may still be misspecified (see Section~\ref{sec:simulation} for an illustration).
Under such misspecification -- either in the reward or preference model -- existing RLHF algorithms can yield suboptimal policies.

To address these model misspecifications, we propose a robust fine-tuning framework to improve the sample efficiency of existing reward-based RLHF algorithms under model misspecification. Unlike previous approaches that aim to extend the BT model \cite[e.g.,][]{fan2024covariate,zhang2024general}, we focus on scenarios where this model is inherently misspecified. Our approach is based on the insight that the reference policy, which generates the responses to be labeled, is typically known or it can be well-specified in practice. We leverage this information to enhance the sample efficiency of the estimated reward and policy under model misspecification.

\begin{figure}[t]
    \centering
    \includegraphics[width=1.0\linewidth]{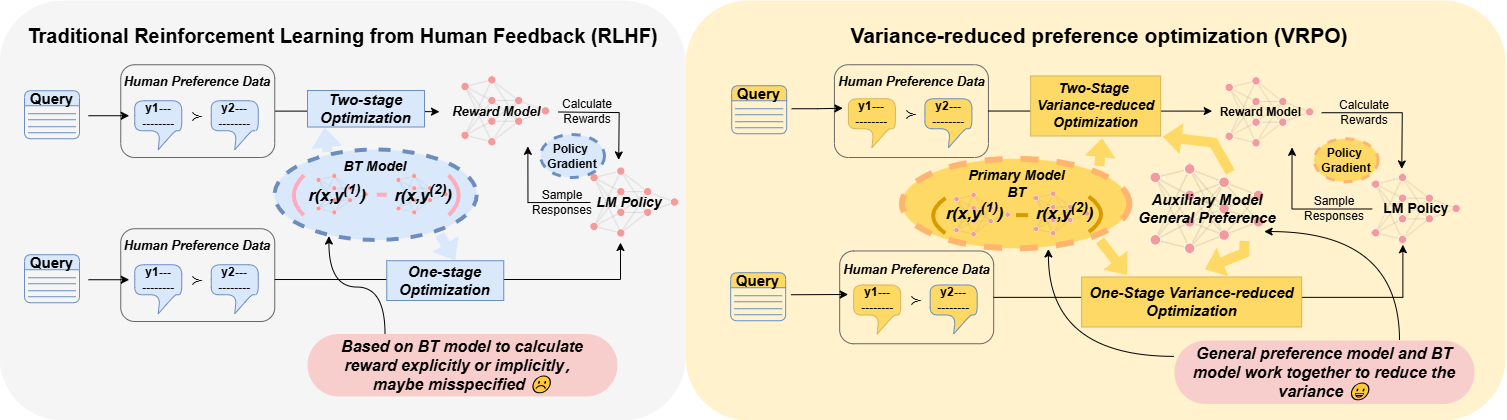}
    \caption{\textbf{VRPO incorporates an auxiliary preference model to reduce the variance of the estimated primary model.} \textbf{Left:} The classic one-stage and two-stage optimization schemes in RLHF. Both approaches require fitting a reward model, either explicitly or implicitly, which may lead to model misspecification. \textbf{Right:} In contrast, VRPO employs an auxiliary reward-free preference model to better capture human preferences. It works jointly with the primary model for variance reduction and policy improvement.}
    \label{fig:workflow}
\end{figure}

\noindent \textbf{Our contribution:}
\begin{enumerate}[leftmargin=*]
    \item We propose variance-reduced preference optimization (VRPO), a flexible pipeline applicable to a variety of existing RLHF algorithms to enhance their sample efficiency under reward or preference model misspecification (see Figure \ref{fig:workflow} for an overview and Section \ref{sec:methodology} for implementation details).
    
    \item We conduct comprehensive numerical experiments on several LLM datasets to demonstrate the superior performance of our algorithm. In particular, on the Anthropic Helpful and Harmless (HH) dataset \citep{bai2022training}, we observe that 53-98\% of the responses generated by our estimated policy are preferred over those produced by 
    \textbf{nine} baselines (see Table \ref{tab:vrpo-variants-temp}).

    \item We rigorously establish the statistical properties of VRPO (see Table \ref{tab:summary} for a summary). Specifically, we prove that when the reward or preference model is misspecified, compared to baseline RLHF algorithms, our method reduces both the variance and mean squared error (MSE) of the estimated parameters (see Theorem \ref{thm:lower MSE of theta}) as well as the suboptimality gap of the resulting policy (see Theorem \ref{thm:suboptimality control}).
\end{enumerate}

\begin{table}[t]
\begin{center}
\begin{small}
\begin{sc}
\begin{tabular}{lccc}
\toprule
\multirow{2}{*}{Model Setting}  &       variance of & MSE of  & suboptimality   \\
&estimator &estimator & gap \\ \midrule
Misspecified &  \tikz{\draw[darkgreen, -latex] (0,-0.6) -- (0,-0.9);}    &\quad\tikz{\draw[darkgreen, -latex] (0,-0.6) -- (0,-0.9);}& \quad\tikz{\draw[darkgreen, -latex] (0,-0.6) -- (0,-0.9);}\\
Correctly specified     &  \parbox[c]{1em}{\tikz{\draw[magenta, -latex] (0,0.2) -- (0.3,0.2);}}        & \quad \parbox[c]{1em}{\tikz{\draw[magenta, -latex] (0,0.2) -- (0.3,0.2);}} &\quad \parbox[c]{1em}{\tikz{\draw[magenta, -latex] (0,0.2) -- (0.3,0.2);}} \\
\bottomrule
\end{tabular}
\end{sc}
\end{small}
\end{center}
\caption{Variance, MSE, and suboptimality gap: proposed VRPO vs. existing RLHF algorithms, where 
\protect\parbox[c]{1em}{\protect\hspace*{2pt}\protect\tikz{\protect\draw[darkgreen, -latex] (0,-0.6) -- (0,-0.9);}}\hspace*{-4pt} represents a decrease and 
\protect\parbox[c]{1em}{\protect\tikz{\protect\draw[magenta, -latex] (0,0.2) -- (0.3,0.2);}} indicates no asymptotic difference.}\label{tab:summary}
\vskip -0.1in
\end{table}

The rest of the paper is organized as follows. We review 
the related work in Section \ref{sec:relatedworks}, introduce the RLHF pipeline in Section \ref{sec:preliminary} and presents the proposed VRPO in Section \ref{sec:methodology}. Sections \ref{sec:experiments} and \ref{sec:theory} are devoted to the finite-sample and theoretical performance of VRPO, respectively. Finally, we conclude our paper in Section \ref{sec:conclusion}. 

\section{Related Work}\label{sec:relatedworks}
This section is organized as follows. We first review existing RLHF algorithms, which can be roughly classified into two categories: reward-based and preference-based. We next discuss a line of robust RLHF algorithms. Finally, we highlight the connection between our proposed approach and semi-supervised learning.

\textbf{Reward-based RLHF.} \citet{christiano2017deep} propose a deep RLHF algorithm that leverages the BT model to learn a reward function from human preference for parameter fine-tuning. They demonstrate the potential of this approach in non-language-based tasks, including robotics and Atari games. Recently, this reward-based approach has gained substantial attention in the development of LLMs, leading to two different fine-tuning approaches in leveraging the reward model.

The first approach employs the reward model explicitly, and we refer to it as \ul{\emph{two-stage optimization}}. Algorithms based on this approach use two stages: they first train a reward model to align with human preferences, and then they utilize RL algorithms such as proximal policy optimization \citep[PPO,][]{schulman2017proximal} to compute the optimal policy based on the estimated reward \citep{ziegler2019fine, bakker2022fine, ouyang2022training, li2024remax, wu2024pairwise, zhang2024cppo, sun2025rethinking}. The second approach uses the reward model in an implicit manner. We refer to this approach as \ul{\emph{one-stage optimization}}. This method is based on the insight that the reward can be derived in closed form using the optimal policy \citep{rafailov2023dpo}. One-stage optimization algorithms parameterize the reward based on the optimal policy, estimating the optimal policy in a single step \citep{rafailov2023dpo, zhao2023calibrating,  hong2024orpo, liu2024statistical, ramesh2024group, shao2024deepseekmath, tang2024gpo, xu2024contrastive, xiao2025cal}. 

Extensions of both approaches have been proposed, including those that use other models than BT for reward-based preferences \citep{tang2024gpo, wang2024beyond, zhang2024general, xiao2025on} and those that integrate active learning to enhance reward training \citep{das2024active, liu2024dual, mukherjee2024optimal}. %and those that handle diverse human feedback \citep{zhong2024provable, li2025preserving, wang2025mpo}. 
Various studies investigated the theoretical properties of these algorithms, including the error bounds of the estimated parameters, the suboptimality gap of the estimated policy, and the alignment objectives of these 
methods \citep[see e.g.,][]{zhu2023principle, nika2024reward, rafailov2024from, xiong2024iplhf,aminian2025theoretical,luo2025learning, vojnovic2025alignment, shi2025understanding}.

\textbf{Preference-based RLHF.} In contrast to reward-based RLHF, preference-based RLHF algorithms search for the optimal policy without modeling the reward. They shift the objective of RLHF from maximizing rewards to optimizing preferences \citep{azar2024general}. Many of these algorithms operate within the Nash learning framework \citep{munos2024nash}. This framework casts policy optimization as a two-player constant-sum game, where the optimal policy is given by the Nash equilibrium \citep{calandriello2024ipo,swamy2024minimaximalist, wu2024self, ye2024online, liu2025statistical, shi2025fundamental, zhang2025iterative}. Beyond Nash learning, several other preference-based algorithms have been proposed, utilizing Bayesian modeling \citep{wang2023aligning}, preference embeddings \citep{hejna2024contrastive}, contrastive learning \citep{hong2024adaptive}, general preference modeling  \citep{zhang2024general} and doubly robust methods \citep{xu2025doubly}.

\textbf{Robust methods for RLHF.} There is growing interest in developing robust RLHF algorithms. 
One line of research focuses on settings where human feedback is corrupted or biased due to subjective judgment \citep{mitchellnote, bukharin2024robust, chowdhury2024provably, wu2024towards}. Another line investigates heterogeneous feedback collected from multiple teachers \citep{hao2023leveraging, freedman2023active, lee2024low, ramesh2024group, zhong2024provable, li2025preserving, wang2025mpo}. Lastly, some robust algorithms are tailored for settings with distribution shifts where deployment prompts can differ substantially from those encountered during training \citep{mandal2025distributionally}. In contrast to these approaches, our robust method differs in objective: we aim to improve the performance of reward-based RLHF algorithms when the reward-based preference model may be misspecified.

\textbf{Semi-supervised learning}. Semi-supervised learning leverages the rich information contained in unlabeled data to enhance supervised learning based solely on labeled data. Our proposal is closely related to a line of work that draws on semiparametric statistics \citep[see e.g.,][]{tsiatis2006semiparametric} for variance-reduced estimation. This idea has been applied to a variety of problems, including semi-supervised linear regression \citep{chakrabortty2018efficient}, mean estimation \citep{zhang2019semi, zhang2022high}, more general objective functions \citep{angelopoulos2023prediction}, settings with covariate shift \citep{zhu2023doubly}, and RL \citep{zhu2025semi}. 

Different from the aforementioned work, we do not develop new semi-supervised learning algorithms. Rather, we adapt ideas from this literature for LLM fine-tuning. Our key observation is that when the reference policy is known, one can generate an abundant amount of responses to each query -- albeit without the corresponding human preference labels. This naturally leads to a semi-supervised learning setting, making these methods well-suited for LLM fine-tuning. To the best of our knowledge, no prior work has leveraged semi-supervised learning in this way for LLM fine-tuning.

\section{Reinforcement Learning from Human Feedback Preliminaries}\label{sec:preliminary}
This section reviews the RLHF pipeline. We first introduce the data generating process (DGP) for LLM fine-tuning. We next introduce the model specification and parameter estimation procedure. 

\textbf{The DGP.} This paper focuses on fine-tuning LLMs using a dataset $\mathcal{D}$, consisting of $n$ tuples of the form $(X, Y^{(1)}, Y^{(2)}, Z)$. These data are generated as follows. At each time step, the LLM receives a \textit{prompt} $X$. Then, the model generates two different \textit{responses} $(Y^{(1)}, Y^{(2)})$ according to a reference policy $\pi_{\textrm{ref}}$ so that $Y^{(1)},Y^{(2)}\sim \pi_{\textrm{ref}}(\ \cdot \ |X)$. The prompt and responses $(X, Y^{(1)}, Y^{(2)})$ are then presented to a human expert, and $Z=\mathbb{I}(Y^{(2)}\succ Y^{(1)})$ represents the resulting \textit{human feedback}. Here, $Y^{(2)}\succ Y^{(1)}$ denotes that the second response $Y^{(2)}$ is preferred, and $\mathbb{I}$ denotes the indicator function.

We primarily focus on settings with a known or well-specified $\pi_{\textrm{ref}}$\footnote{``Well-specified'' means the specified $\pi_{\textrm{ref}}$ is very close to the ground truth.}. This assumption is automatically satisfied in the aforementioned scenario, where the responses are generated by the LLM we wish to fine-tune, either using the pre-trained policy or the SFT policy. When using external dataset for fine-tuning, this assumption can also be satisfied if well-trained reference policies are available, as is the case with the HH dataset (see Section \ref{sec:experiments} for details).

\textbf{The model.} There are two crucial components under the RLHF framework:
(i) A reward function $r^*(X,Y)$ that measures the favorability of a response $Y$ to a prompt $X$; and (ii) a preference function $p^*(X, Y^{(1)}, Y^{(2)}) = \mathbb{P}(Y^{(2)}\succ Y^{(1)}|X)$ that determines the probability of $Y^{(2)}$ being preferred over $Y^{(1)}$ given $X$. Our goal is to learn an optimal policy to generate the best response for each prompt, maximizing the expected reward. However, as mentioned in the introduction, it remains highly challenging to ask human experts to directly provide numeric values to mimic the oracle reward $r^*$. Instead, RLHF learns the optimal policy through the preference model. Existing reward-based RLHF algorithms assume the preference $p^*$ can be computed based on the difference between two rewards, such that
\begin{equation}\label{eqn:peferencereward}
    p^*(x,y^{(1)},y^{(2)})=\sigma(r^*(y^{(2)},x)-r^*(y^{(1)},x)),
\end{equation}
for some activation function $\sigma: \mathbb{R} \to [0,1]$. For example, in the BT model, $\sigma$ is set to the sigmoid function. 

Under a reward-based preference model assumption as in Equation \eqref{eqn:peferencereward}, these algorithms approximate $r^*$ using a parametric model $r_{\theta}$, indexed by parameters $\theta$. This approximation, in turn, gives rise to the parametric preference model $p_{\theta}(x,y^{(1)},y^{(2)})=\sigma(r_{\theta}(y^{(2)},x)-r_{\theta}(y^{(1)},x))$ for approximating the true preference function $p^*$.

\textbf{Parameter estimation.} The estimator $\widehat{\theta}$ is computed by minimizing the empirical risk $\mathcal{L}(\theta)=\mathbb{E}_n [\ell(X,Y^{(1)},Y^{(2)},Z;\theta)]$ where $\ell$ denotes certain loss function, e.g., the cross-entropy loss given by 
\begin{eqnarray}\label{eq:cross-entropy}
    \ell(x,y^{(1)},y^{(2)},z;\theta) = - z\log(p_\theta(x,y^{(1)},y^{(2)})) - (1-z)\log(1-p_\theta(x,y^{(1)},y^{(2)})),
\end{eqnarray}
and $\mathbb{E}_n$ denotes the empirical average over the dataset $\mathcal{D}$. This leads to the estimated reward $r_{\widehat{\theta}}$, which can then be used to derive the estimated optimal policy, for example, through PPO. 

We remark that the consistency of the estimated policy relies on the correct specification of both the reward and preference models. However, model misspecification can lead to sub-optimal policies. There are two potential sources of misspecification. First, misspecification occurs when Equation \eqref{eqn:peferencereward} fails to hold, due to an incorrect activation function $\sigma$, or due to the fact that $p^*$ cannot be represented as a function of the difference between two rewards. Second, misspecification occurs if $r^*$ cannot be represented by the chosen parametric model $r_{\theta}$ for any $\theta$. Our proposal, detailed in the next section, aims to improve policy learning under these misspecifications. We empirically illustrate both types of misspecifications in Sections \ref{sec:simulation} and \ref{sec:real-data}, respectively. 

\section{Variance-Reduced Preference Optimization}\label{sec:methodology}
In this section, we present our proposed algorithm and demonstrate how it can be integrated into existing reward-based RLHF algorithms to improve their sample efficiency. Our proposal estimates two models for the preference function $p^*$: (i) A primary, simpler reward-based model $p_{\theta}$, as in existing algorithms; and (ii) an auxiliary, more complex preference model $p_{\eta}$, designed to enhance the accuracy of the primary model. We remark that the auxiliary model is trained separately and ahead of the primary model, for example, by minimizing the empirical cross-entropy loss (see Equation \eqref{eq:cross-entropy}). It differs from the primary model in that it can either be non-reward-based, utilizing more general preference functions \citep[e.g.,][]{zhang2024general} to mitigate the misspecification of $p^*$, or remain reward-based but involving a larger number of parameters to alleviate the misspecification of $r^*$. By design, this more complex model is expected to be more robust to model misspecification compared to the primary model.

After training the auxiliary model, we estimate the parameter $\theta$ in the primary model by minimizing the following loss function
\begin{equation}\label{eq:modified-loss}
\begin{split}
    \widetilde{\mathcal{L}}(\theta) &=  \mathbb{E}_n\Big[ \ell(X, Y^{(1)}, Y^{(2)}, Z; \theta) - \sum\limits_{u=0}^1\ell(X, Y^{(1)}, Y^{(2)}, u; \theta))p_\eta(X, Y^{(1)}, Y^{(2)}, u) \\
    &\qquad\qquad + \sum\limits_{u=0}^1 \mathbb{E}_{y^{(1)*}, y^{(2)*}\sim\pi_{\textrm{ref}}(\ \cdot \ |X)}\ell(X, y^{(1)*}, y^{(2)*}, u; \theta)p_\eta(X, y^{(1)*},  y^{(2)*}, u)\Big],
\end{split}
\end{equation}
By definition, the above objective function consists of three terms. The first term\\$ \mathbb{E}_n [\ell(X, Y^{(1)}, Y^{(2)}, Z; \theta)]$ is the loss function of existing reward-based RLHF algorithms. The second and third terms correspond to loss functions constructed using data generated by the auxiliary model $p_{\eta}$ and the reference policy $\pi_{\textrm{ref}}$. The primary roles of the last two terms are to reduce the variance while ensuring unbiasedness. Specifically, if $\pi_{\textrm{ref}}$ is correctly specified, the last two terms share the same mean, and their expectations cancel out. Meanwhile, when $p_\eta$ is close to the oracle preference function $p^*$, including the last two terms reduces the variance of the objective function. 

Our modified loss function yields the following two properties, which we will formally establish in Section \ref{sec:theory} and empirically validate in Section \ref{sec:experiments}:
\begin{enumerate}[leftmargin=*]
    \item The first is the \textbf{variance reduction} property, which is the primary benefit of our proposal. Specifically, when either the reward or preference model (e.g., BT) is misspecified, the variance reduction in the loss function will lead to \ul{\emph{a reduction in the variance and MSE of the resulting estimator $\widetilde{\theta}$, as well as the suboptimality gap of the estimated policy}}; 
    \item The second is the \textbf{doubly robust} property, which is secondary and a byproduct that arises when both the reward and preference models are correctly specified. It guarantees consistency when \ul{\emph{either the reference policy $\pi_{\textrm{ref}}$ or the auxiliary model $p_{\eta}$ is correctly specified}}. 
\end{enumerate}
%We will formally show in Section \ref{sec:theory} that 
To conclude this section, we illustrate how our proposal can be applied to both two-stage and one-stage optimization described in Section \ref{sec:relatedworks}. 

\textbf{Two-stage optimization}. In two-stage optimization, we directly model the reward as $r_{\theta}$ and specify an activation function $\sigma$, leading to the preference model $p_{\theta}(x,y^{(1)},y^{(2)})=\sigma(r_{\theta}(y^{(2)},x)-r_{\theta}(y^{(1)},x))$. We next choose a proper loss function for $\ell$, such as the cross-entropy loss or hinge loss, and compute the estimated parameter $\widetilde{\theta}$ by minimizing \eqref{eq:modified-loss}. Finally, we apply a standard RL algorithm (e.g., PPO) to estimate the optimal policy by solving
\begin{equation}\label{eqn:rewardpolicy}
    \arg\max_\pi \mathbb{E}_n \Big[\mathbb{E}_{y\sim\pi(\ \cdot \ |X)}[r_{\widetilde{\theta}}(X, y)] - \beta \cdot\text{KL}(\pi(\ \cdot \ |X)||\pi_{\textup{ref}}(\ \cdot \ |X))\Big],
\end{equation}
for some tuning parameter $\beta>0$, 
where $\text{KL}$ denotes the Kullback-Leibler divergence.

\textbf{One-stage optimization}. In one-stage optimization, we parameterize the optimal policy via $\pi_\theta$. Solving \eqref{eqn:rewardpolicy} yields the following close-form expression for the reward model  %given by $\beta[\log \pi_{\theta}(y|x)-\log \pi_{\textrm{ref}}(y|x)]$. 
\begin{eqnarray}\label{eqn:rewardpolicy2}
    r_{\theta}(x,y)=\beta \log \left(\frac{\pi_{\theta}(y|x)}{\pi_{\textrm{ref}}(y|x)}\right)-C(x),
\end{eqnarray}
where $C(x)$ denotes some baseline term that is independent of the response. We next employ this reward to model the preference, and minimize the resulting modified loss function \eqref{eq:modified-loss} for estimating the policy parameter $\theta$.

\section{Experiments}\label{sec:experiments}

\begin{table}[t]
\centering
\caption{Summary of experimental settings. Here,  {\color{ForestGreen}\ding{51}} indicates that the model is likely correctly specified, {\color{red}\ding{55}} indicates that the model is misspecified, \textbf{?} denotes cases where it is unclear whether the model is correct, and {\color{ForestGreen}\ding{51}} / {\color{red}\ding{55}} indicates cases that include both correctly specified and misspecified settings.}\label{tab:experiments_summary}
\renewcommand{\arraystretch}{1.2}
\begin{tabular}{c l l c c c}
\toprule
Section & Task & Data & \makecell{Preference\\model} & \makecell{Reward\\model} & \makecell{Reference\\policy} \\
\midrule
\ref{sec:simulation} & Sentiment generation & IMDb
& {\color{ForestGreen}\ding{51}} 
& {\color{red}\ding{55}} 
& {\color{ForestGreen}\ding{51}} / {\color{red}\ding{55}} \\
\midrule
\ref{sec:real-data} & Dialogue & Anthropic HH
& {\color{red}\ding{55}} 
& \textbf{?}
& {\color{ForestGreen}\ding{51}} \\
\midrule
\ref{sec:TLDR} & Summarization & TL;DR
& \textbf{?}
& \textbf{?}
& {\color{red}\ding{55}} \\
\bottomrule
\end{tabular}
\end{table}

In this section, we empirically 
evaluate the performance of the proposed VRPO in three complex RLHF tasks, including sentiment generation (Section \ref{sec:simulation}), dialogue (Section \ref{sec:real-data}) and summarization (Section \ref{sec:TLDR}). These tasks are widely used in the literature for evaluation \citep{rafailov2023dpo,wu2024pairwise,lu2025discovering, xiao2025cal}. Each task targets a difference source of model misspecification: (i) the reward model, (ii) the preference model, and (iii) the reference policy. Table~\ref{tab:experiments_summary} summarizes the experimental settings for all three tasks. 

\textbf{Implementation}. Our implementation parallels the training pipeline of direct preference optimization \citep[DPO,][]{rafailov2023dpo}, while adopting the loss function defined in~\eqref{eq:modified-loss}. Computing this loss requires (a) a reference policy $\pi_{\textup{ref}}$; (b) an approximation of the expectation term in~\eqref{eq:modified-loss}; (c) a preference model $p_\eta$. For (a), except in the summarization task, we trained an SFT policy via token-wise supervised fine-tuning \citep{rafailov2023dpo, wu2024pairwise} as $\pi_{\textup{ref}}$ . For (b), we approximate the expectation term using Monte Carlo by sampling a pair of responses, $y^{(1)}$ and $y^{(2)}$, from the reference policy~$\pi_{\textup{ref}}$ given the prompt~$x$. For (c), we estimate $p_{\eta}$ using the BT model in Section \ref{sec:simulation} and the general preference model \citep{zhang2024general} in Sections \ref{sec:real-data} and \ref{sec:TLDR}. Further implementation details are provided in Section S.2 of the Supplementary Material A. In the following, we present the main results of the proposed VRPO under one-stage optimization (more results in Section S.3.1 of the Supplementary Material A), while results based on the two-stage optimization are presented in  Section S.3.2 of the Supplementary Material A.

\textbf{Summary of findings}. Before presenting the detailed results, we briefly summarize our main empirical observations:
\begin{itemize}[leftmargin=*]
    \item Section~\ref{sec:simulation} demonstrates that VRPO achieves superior performance than DPO when the reward model is misspecified, and validates its doubly robust property through simulation.
    \item Section~\ref{sec:real-data} shows that VRPO consistently outperforms \textbf{9} baseline algorithms, including DPO and its robust variants, when the preference model is misspecified.
    \item Section%s~\ref{sec:simulation} and
    ~\ref{sec:TLDR} offers a sensitivity analysis of VRPO with respect to the reference policy. Even when the reference policy is misspecified, VRPO maintains better performance than SFT and DPO. 
\end{itemize}

\subsection{Synthetic Data Analysis under Reward Model Misspecification}\label{sec:simulation}
We first conduct a simulation study that generates human preferences following the BT model to validate the variance reduction and doubly robust properties of VRPO.

\textbf{Task, dataset and evaluation criterion}. We consider the IMDb dataset \citep{Andrew_2011_Learningwordvectors}, which contains 50,000 movie reviews evenly balanced between positive and negative sentiment. The specific task we consider is \textit{controlled sentiment generation}, in which the objective is to steer a pretrained language model to produce positive movie reviews. Specifically, we use 25,000 prefixes of movie review sentences from the training set as prompts $X$. For each prompt $X$, two candidate responses $Y^{(1)}$ and  $Y^{(2)}$ are generated by an SFT model trained on the same IMDb training data. Rewards for each response are annotated based on a pre-trained sentiment classifier \texttt{siebert/sentiment-roberta-large-english} \citep{hartmann2023}, and preference labels between responses are subsequently sampled using the BT model so that the reward-based preference condition \eqref{eqn:peferencereward} is satisfied.

 Because the preferences in this task are synthetically generated from a known reward function, model performance can be directly assessed by the expected reward of generated responses on the remaining 25,000-sample test set.

\begin{figure}[t]
        \centering
        \begin{minipage}[t]{0.32\textwidth}
            \centering
            \includegraphics[width=\linewidth]{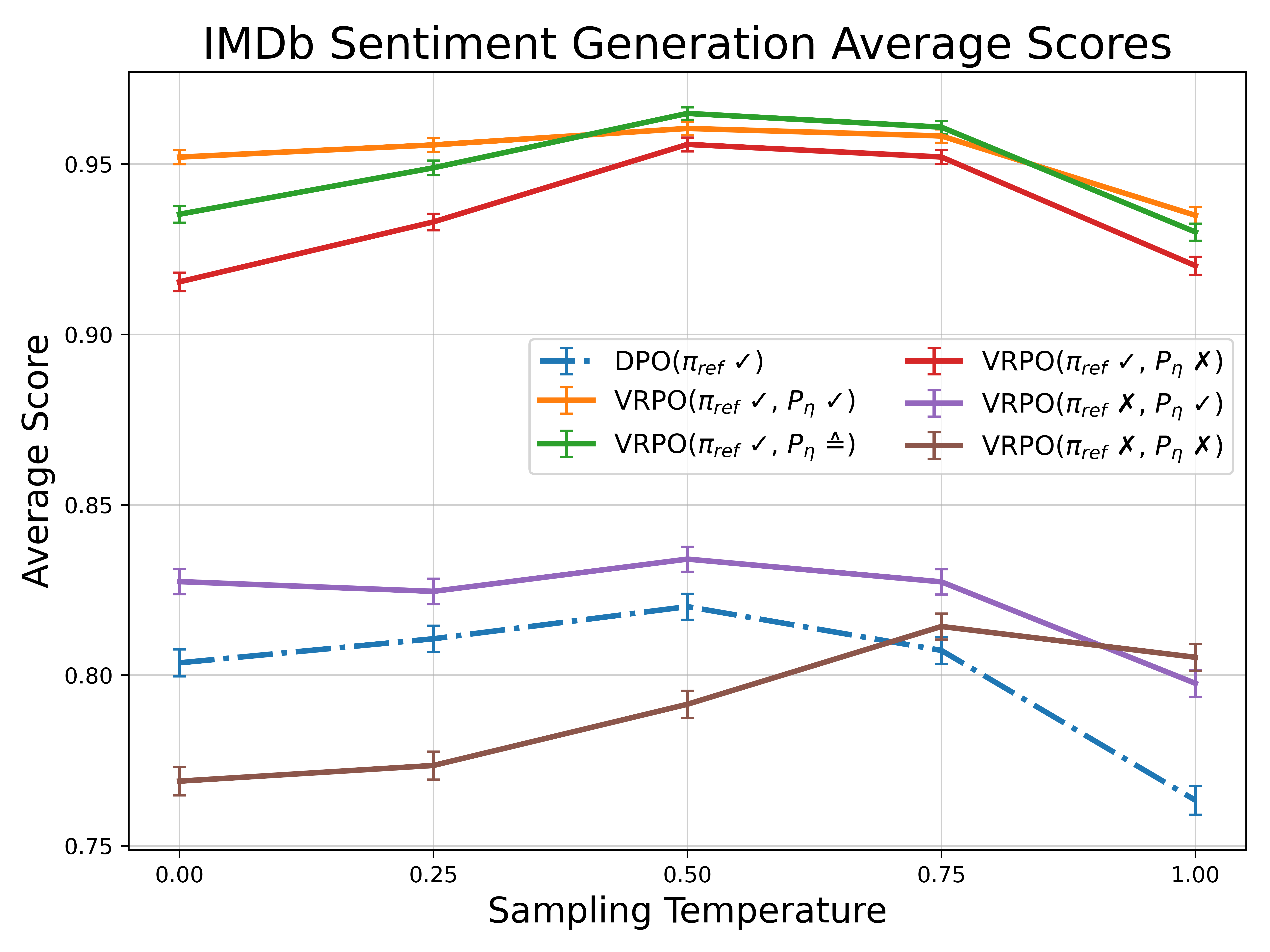}
        \end{minipage}
        \hfill
        \begin{minipage}[t]{0.32\textwidth}
            \centering
            \includegraphics[width=\linewidth]{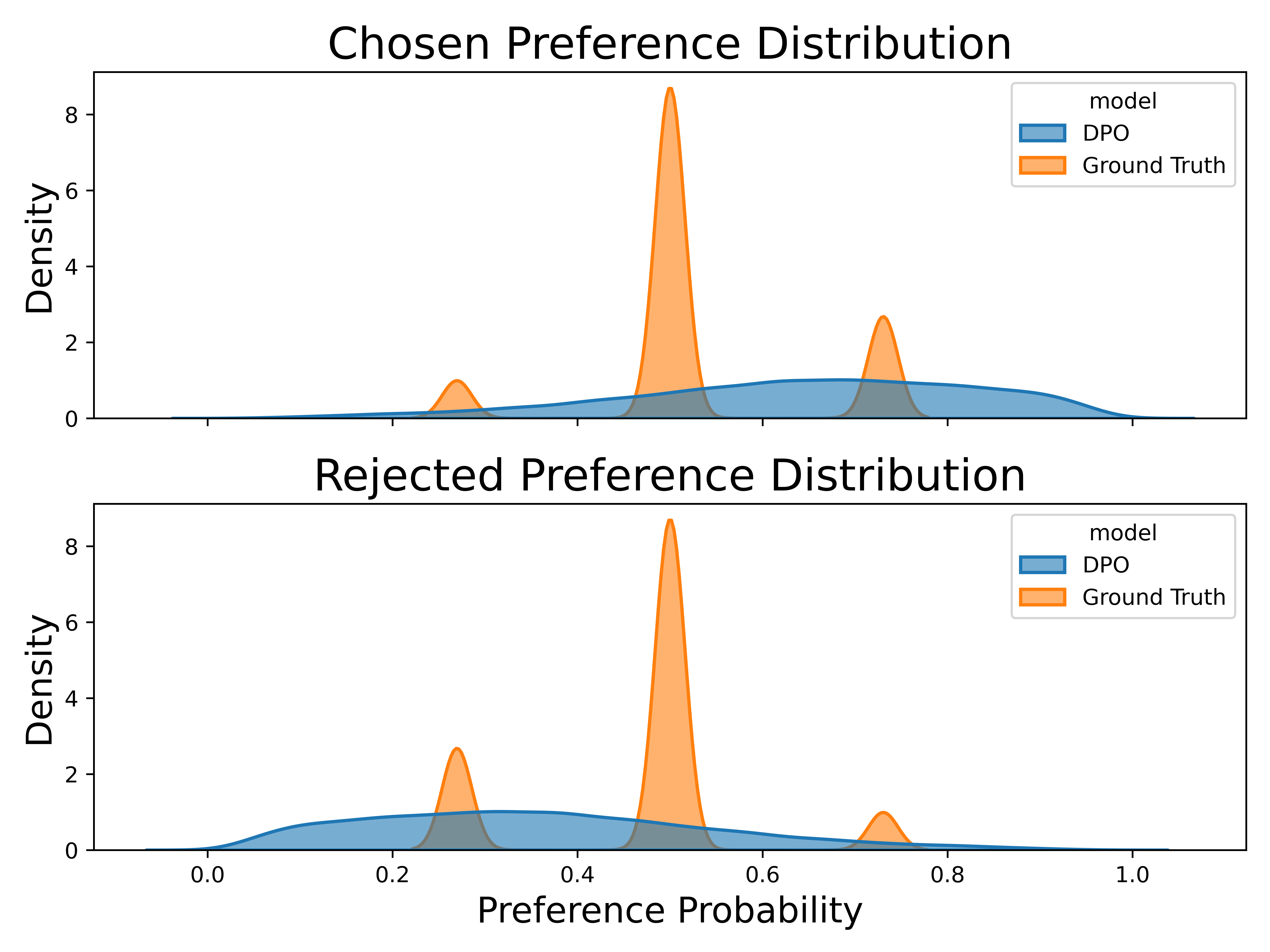}
        \end{minipage}
        \hfill
        \begin{minipage}[t]{0.32\textwidth}
            \centering
            \includegraphics[width=\linewidth]{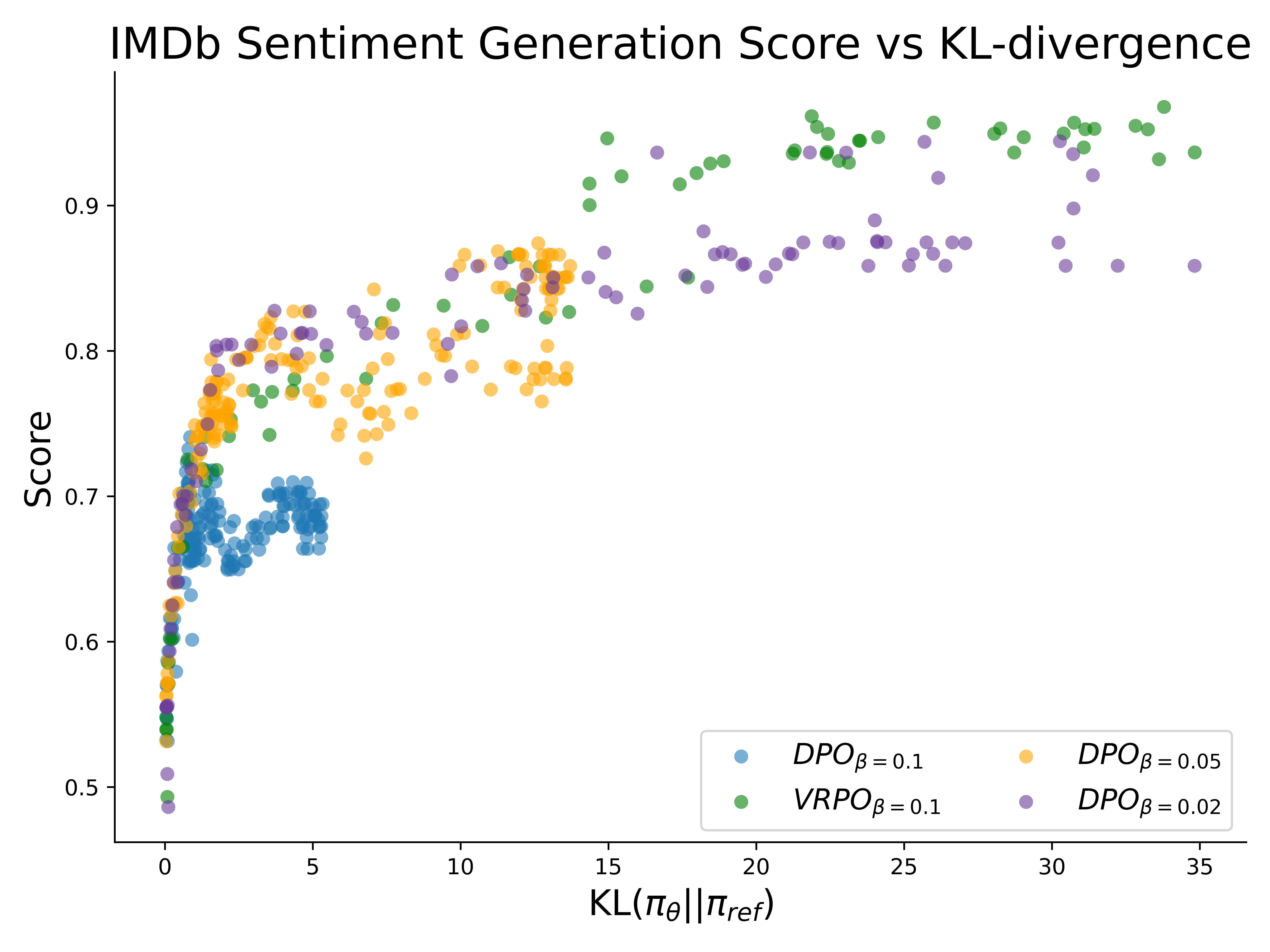}
        \end{minipage}
    \caption{Comparisons in IMDb dataset. \textbf{Left} panel represents the expected reward in different VRPO setting compared to DPO, for example $(\pi_{ref}$ \ding{51}, $P_{\eta}$ \ding{55}$)$ means the reference model is correctly specified and the preference model is misspecified, and $P_{\eta}$$\boldsymbol{\hat{=}}$ means the preference model is estimated, demonstrating the robustness of our method. \textbf{Middle} plane illustrates the difference in preference probability distributions between the ground truth and the DPO estimation for both the chosen and rejected responses. \textbf{Right} The panel reports the expected reward versus KL-divergence for VRPO with $\beta = 0.1$ and DPO with $\beta \in \{0.02, 0.05,0.1\}$.}
    \label{fig:kl-reward}
\end{figure}

\textbf{Result I: variance reduction property.} 
We implement DPO (blue line) and a variant of VRPO (green line) that uses DPO's estimated reward to construct the auxiliary model, and report their expected rewards in the left panel of Figure~\ref{fig:kl-reward}. Recall that the preference model is correctly specified in this task. As summarized in Table~\ref{tab:summary}, when the reward model is misspecified, VRPO is theoretically expected to perform comparably to DPO. However, it achieves superior empirical performance. We suspect this improvement is due to DPO's reward model misspecification. Specifically, using the optimal policy in \eqref{eqn:rewardpolicy2} does not perfectly recover the true reward, resulting in a misspecified reward model. 
The middle panel of Figure~\ref{fig:kl-reward} provides supporting evidence for such reward model misspecification, which shows that DPO's fitted preference distribution drived from its estimated reward deviates substantially from the true distribution obtained using the oracle reward. 

The primary goal in RLHF is to maximize the reward while constraining the policy’s divergence from the reference policy $\pi_{\textup{ref}}$. Ideally, a superior RLHF algorithm shall achieve higher expected reward at each fixed KL-divergence level between the learned policy and $\pi_{\textup{ref}}$. It can be seen from the right panel of Figure~\ref{fig:kl-reward} that, with the tuning parameter $\beta$ set to 0.1 (see \eqref{eqn:rewardpolicy} and \eqref{eqn:rewardpolicy2}), VRPO generally attains higher rewards than DPO at each fixed KL-divergence level. Since DPO with $\beta=0.1$ struggles to learn substantially different policies from $\pi_{\textup{ref}}$, we additionally evaluate DPO by setting $\beta$ to 0.05 and 0.02. In this extended comparison,  VRPO maintains its advantage, achieving higher rewards at comparable or lower KL-divergence levels.

All the aforementioned results illustrate VRPO's variance reduction property: even under correct preference model specification (e.g., BT), VRPO can still improve existing algorithms if their reward model is misspecified. 

\textbf{Result II: doubly robust property.} By construction, our synthetic environment satisfies the reward-based preference condition. %Under correct (or approximately correct) reward model specification, the setup closely mirrors the ideal setting without model-misspecification. This allows us 
To empirically validate the doubly robust property established in Theorem~\ref{thm:doubly-robust property}, %Specifically, 
we implement four versions of VRPO, depending on whether $p_{\eta}$ and $\pi_{\textrm{ref}}$ are correctly specified, and report their expected rewards in the left panel of Figure~\ref{fig:kl-reward}. The oracle expected reward is 1. It can be seen that VRPO with both correctly specified $p_{\eta}$ and $\pi_{\textrm{ref}}$ (orange line) achieves the highest expected reward (around 0.95) across most temperatures. When either one model is correctly specified, the expected reward varies from 0.8 to 0.95 (red and purple lines). Finally, the expected reward drops to below 80\% when both models are misspecified (brown line). These results empirically confirm VRPO's doubly robust property. 

As mentioned in Section \ref{sec:methodology}, the validity of the doubly robust property requires that both the preference model and the reward model be correctly specified. However, in this task, the reward model is likely misspecified. Despite this, we still observe the doubly robust property.

\subsection{Real Data Analysis under Preference Model Misspecification}\label{sec:real-data}
We next apply VRPO to a real-world dataset and compare it against nine baseline algorithms to demonstrate its effectiveness. 

\textbf{Task and dataset.} 
We consider the task of \textit{dialogue generation}, where the prompt $X$ ranges from scientific questions to emotional support suggestions, and the objective is to generate responses $Y$ that are factually accurate, emotionally appropriate, and closely aligned with human expectations. Toward that end, we adopt the widely used Anthropic Helpful and Harmless (HH) dataset \citep{bai2022training}, which consists of multi-turn dialogues between a human and an automated assistant. For each dialogue history $X$, two responses $Y^{(1)}$ and $Y^{(2)}$ are provided for the final turn, and a preference label $Z$ is assigned indicating which response better aligns with human judgment. 

This dataset is well-suited for illustrating preference model misspecification. Prior work has observed that this dataset likely contains unknown pairwise noise that is not captured by the BT model \citep{chowdhury2024provably,wu2024pairwise}. Specifically, both papers studied pairwise noise, where preference labels may be flipped in the training data. Their proposals require to specify a tuning parameter $\epsilon$ representing the label-flip probability, and empirical results show that setting $\epsilon=0.1$ yields the best performance on the HH dataset. This suggests that HH contains certain levels of pairwise noise, leading to the violation of the assumed BT model assumption.

\textbf{Baselines algorithms.} We compare against nine baseline algorithms, including (i) SFT; (ii) DPO; (iii) contrastive preference optimization \cite[CPO,][]{xu2024contrastive}; (iv) DPO-hinge \citep{zhao2023calibrating}; (v) identity preference optimization \cite[IPO,][]{azar2024general}; (vi) conservative DPO \cite[cDPO,][]{mitchellnote}; (vii) robust DPO \cite[rDPO,][]{chowdhury2024provably}; (viii) distributionally robust DPO \cite[Dr. DPO,][]{wu2024towards} and (xi) odds ratio preference optimization \cite[ORPO,][]{hong2024orpo}. Among them, (iii)--(viii) are variants of DPO. Specifically, (iii) replaces DPO’s reference policy with a uniform policy to save memory and facilitate computation; (iv) and (v) replace DPO’s cross-entropy loss with hinge and squared losses, respectively; (vi)--(viii) extend DPO to settings with pairwise noise. Finally, (ix) introduces an odds-ratio penalty into the SFT objective to achieve preference alignment.

To implement VRPO, we use DPO-hinge as the base fine-tuning algorithm to construct the modified loss function in Equation~\eqref{eq:modified-loss}.
The use of hinge loss here helps prevent exploding gradients.
%For DPO-hinge, which performs worse than DPO (see Table~\ref{tab:vrpo-variants-temp}), we additionally implement a variant of VRPO using DPO-hinge as the base algorithm. This serves as an ablation study to examine VRPO's performance when integrated with alternative base algorithms.

\textbf{Evaluation criteria}. In the absence of a ground-truth reward function, Human evaluation is widely regarded as the gold standard for assessing the quality and accuracy of LLMs \citep{ouyang2022training, rafailov2023dpo}. Nevertheless, its inherent limitations in scalability and reproducibility motivate the use of automated evaluation methods. In this task, we evaluate the algorithms' performance on both (i) in-distribution data across five sampling temperatures (0.0, 0.25, 0.5, 0.75, 1.0) and (ii) out-of-distribution data, respectively. 

For (i), we follow \citet{rafailov2023dpo, zhang2024general} and employ GPT as the automatic evaluator. This approach is supported by empirical evidence that LLM-based assessments often outperform traditional evaluation methods \citep{chen2023exploring}.
Specifically, for each temperature, we perform pairwise comparisons as follows: for each pair of fine-tuning algorithms, (a) responses are generated on the same testing set of prompts from the HH dataset, (b) preferences are elicited using GPT as the evaluator, and (c) win rates are computed as the proportion of cases in which one method's response is preferred over the other's.

For (ii), we use the  AlpacaEval~2.0 benchmark \citep{dubois2024length}.  AlpacaEval~2.0 is an LLM-based automatic evaluator. Its prompt set is derived from AlpacaFarm \citep{dubois2023alpacafarm}, which contains a broad collection of human-written instructions representing general-purpose tasks beyond HH. For each prompt, both the model-generated response and a reference response produced by the SFT policy are submitted to a GPT-4-Turbo-based annotator, which determines the preferred response. These preferences are then averaged to compute the win rate of each fine-tuning algorithm against SFT.

\textbf{Result I: in-distribution performance.} 
Table~\ref{tab:vrpo-variants-temp} reports the in-distribution win rates of the proposed VRPO against each of the baseline algorithms. It can be seen that VRPO consistently outperforms all baselines across different sampling temperatures, achieving win rates above 50\% in all cases. In particular, responses generated by VRPO are preferred over those generated by standard DPO approximately 77–81\% of the time.

As discussed earlier, the objective function of VRPO is constructed using a hinge loss rather than the cross-entropy loss employed in DPO. The comparison with DPO-hinge thus serves as an ablation study, where VRPO maintains win rates consistently over 80\%. Moreover, Table~\ref{tab:vrpo-variants-temp} shows that DPO-hinge performs worse than the original DPO. These results suggest that VRPO’s performance gains primarily stem from its variance reduction property rather than the specific choice of loss function.

\begin{table*}[t]
\centering
\caption{Win rates of VRPO and DPO variants across different sampling temperatures.}
\label{tab:vrpo-variants-temp}
\begin{tabular}{@{}lccccc@{}}
\hline
 & \multicolumn{1}{c}{0.0} & \multicolumn{1}{c}{0.25} & \multicolumn{1}{c}{0.5} & \multicolumn{1}{c}{0.75} & \multicolumn{1}{c}{1.0} \\
\hline

VRPO v.s.\ Dr.\ DPO & 0.64 & 0.63 & 0.60 & 0.55 & 0.53 \\
VRPO v.s.\ DPO    & 0.79 & 0.81 & 0.81 & 0.80 & 0.78 \\
VRPO v.s.\ rDPO   & 0.72 & 0.70 & 0.68 & 0.64 & 0.59 \\
VRPO v.s.\ cDPO   & 0.77 & 0.81 & 0.79 & 0.76 & 0.68 \\
VRPO v.s.\ IPO    & 0.80 & 0.81 & 0.77 & 0.78 & 0.75 \\
VRPO v.s.\ DPO-hinge & 0.83 & 0.83 & 0.84 & 0.84 & 0.81 \\
VRPO v.s.\ CPO    & 0.95 & 0.94 & 0.90 & 0.87 & 0.85 \\
VRPO v.s.\ ORPO   & 0.96 & 0.95 & 0.98 & 0.95 & 0.94 \\
DPO-hinge v.s.\ DPO  & 0.42 & 0.44 & 0.46 & 0.40 & 0.44 \\

\hline
\end{tabular}
\end{table*}

\textbf{Result II: out-of-distribution performance}. Table~\ref{tab:hh-winrate_alapacaeval} reports the out-of-distribution win rates of each preference optimization algorithm on AlpacaEval~2.0, evaluated against SFT. The two columns summarize the standard win rate and the length-controlled (LC) win rate adjusted by the length of the response. VRPO again attains the highest win rates, achieving 93.79\% (92.95\% LC), compared to 90.93\% (92.16\%) for the strongest competitor, Dr. DPO. Other variants such as CPO, ORPO and IPO perform markedly worse. 

Taken together, these findings show that VRPO not only achieves superior in-distribution performance but also generalizes more effectively to unseen training prompts, demonstrating the benefits of variance reduction in preference-based optimization under preference model misspecification.

\begin{table}[t]
\centering
\caption{Win rate and length-controlled win rate of different algorithms relative to the SFT policy on the HH dataset by AplacaEval 2.0. Higher win rates indicate better performance. The highest one is present in \textbf{bold}.}
\label{tab:hh-winrate_alapacaeval}
\begin{tabular}{@{}lcc@{}}
\hline
Model & LC Win Rate (\%) & Win Rate (\%) \\
\hline
DPO     & 83.90 & 84.09 \\
Dr.\ DPO & 92.16 & 90.93 \\
rDPO    & 86.89 & 85.71 \\
cDPO    & 85.05 & 84.28 \\
CPO     & 73.59 & 71.28 \\
ORPO    & 75.92 & 53.91 \\
IPO     & 78.29 & 78.88 \\
DPO-hinge     & 80.62 & 79.50 \\
VRPO    & \textbf{92.95} & \textbf{93.79} \\
\hline
\end{tabular}
\end{table}

\subsection{Real Data Analysis under Reference Policy Misspecification}\label{sec:TLDR}
Effective variance reduction in VRPO relies on a well-specified behavior policy. To conclude this section, we conduct a sensitivity analysis to examine its performance in settings where the reference policy is misspecified.

\textbf{Task, dataset and evaluation}. We consider the task of \textit{summarization}, which focuses on producing concise yet informative summaries from long-form text inputs. We utilize the TL;DR (“Too Long; Didn’t Read”) dataset \citep{Vlske_2017_TLDRMR}, which contains different Reddit posts with preference annotations collected by \citet{stienon2020learning}.  Here, the prompts $X$ correspond to Reddit posts, often lengthy and diverse in content, and the objective is to generate summaries $Y$ that are both succinct and faithful to the original information. For evaluation, we compare VRPO against SFT and DPO, and conduct head-to-head comparisons to compute pairwise win rates among the three methods.

This dataset is well-suited for illustrating reference policy misspecification. Specifically, we use the public model trained by \href{https://huggingface.co/cleanrl}{\textit{CleanRL}} as the reference policy. However, this SFT model was trained on a filtered version of the dataset, which differs from the full TL;DR dataset used in our experiments \citep{huang2024n+}, leading to a misspecified reference policy. 

\begin{figure}[t]
    \centering
    \begin{minipage}[t]{0.42\textwidth}
        \centering
        \includegraphics[width=\linewidth]{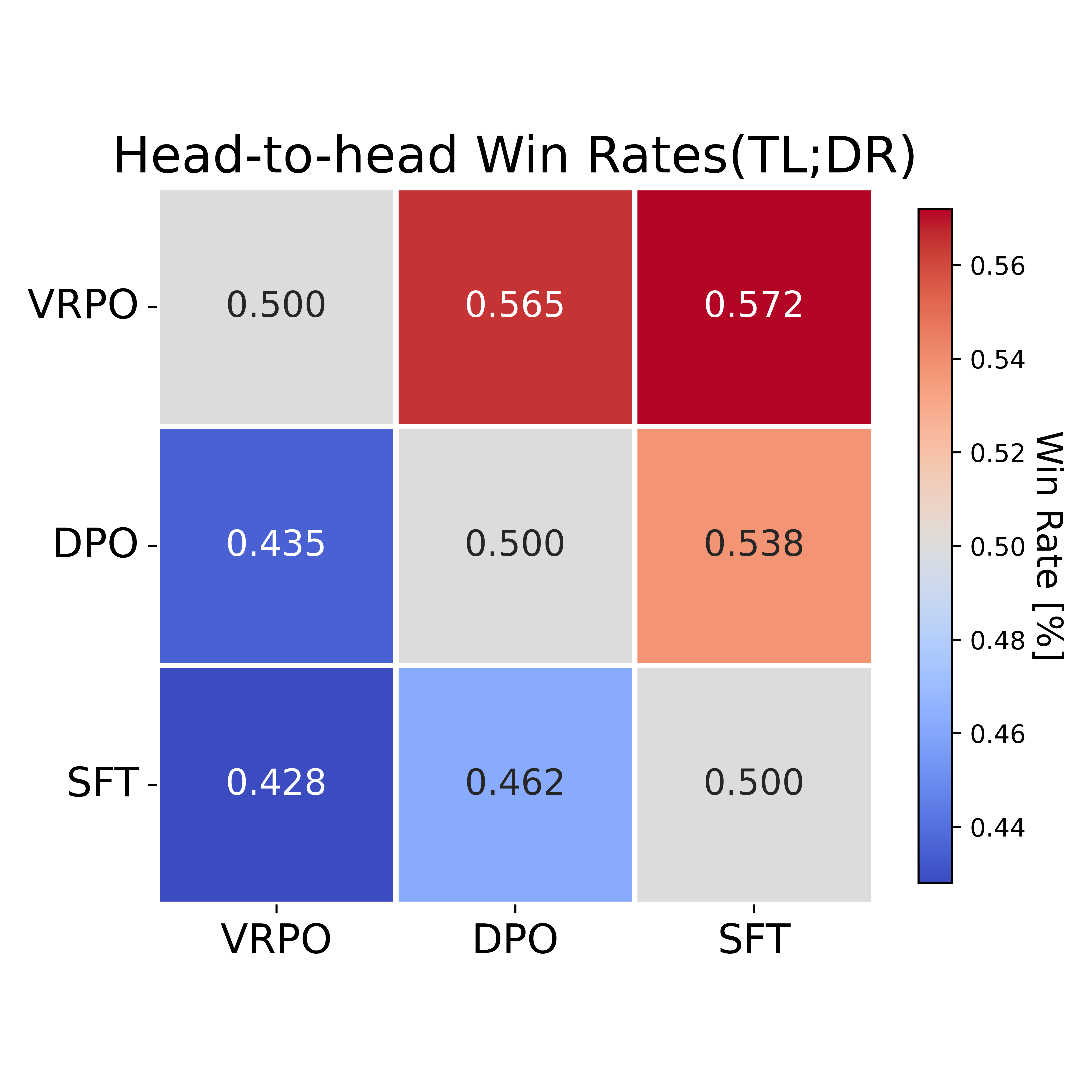}
    \end{minipage}
    \centering
    \begin{minipage}[t]{0.42\textwidth}
        \centering
        \includegraphics[width=\linewidth]{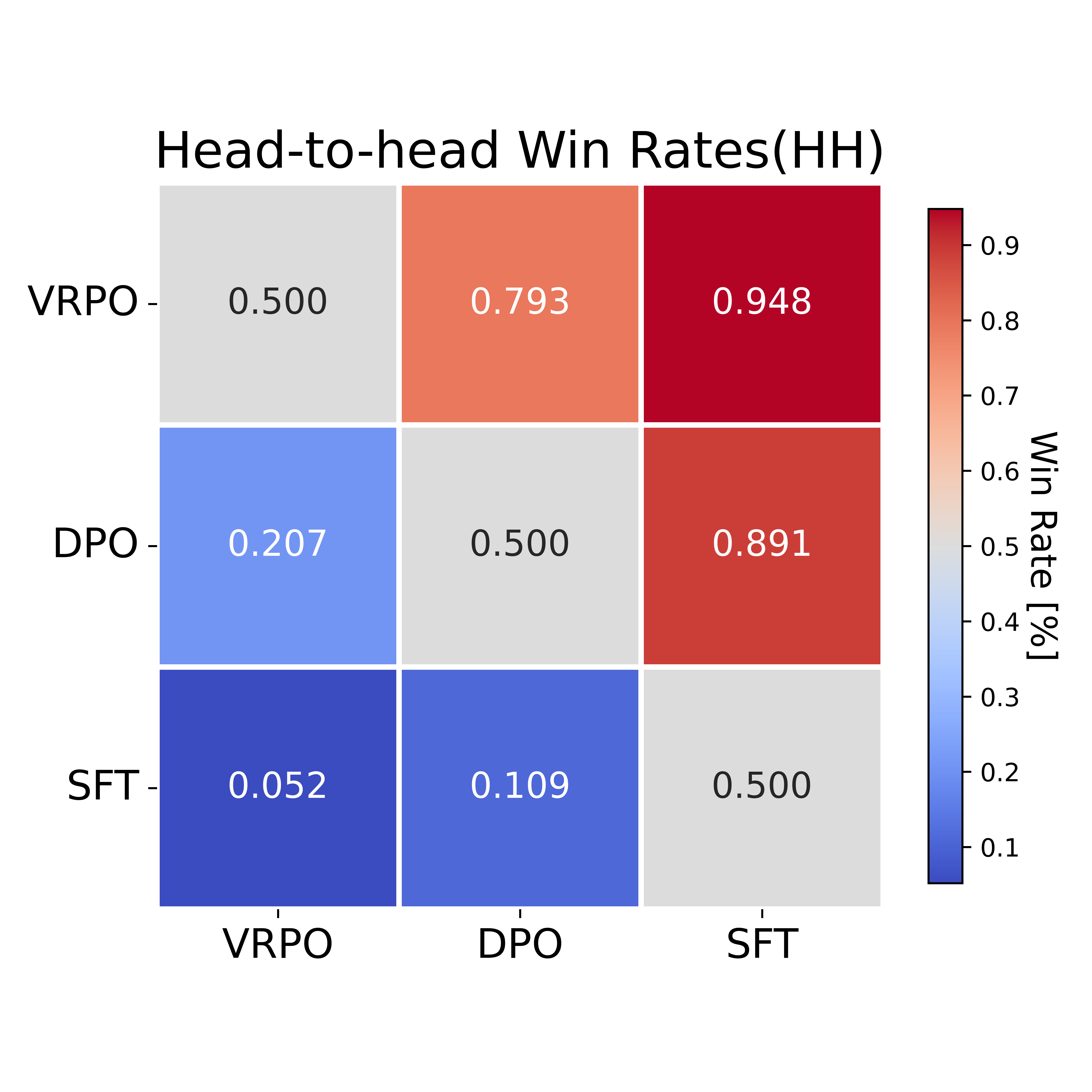}
    \end{minipage}
    \caption{Head-to-head comparisons between {VRPO, DPO, SFT}. Win rates are evaluated by GPT-4o-mini. \textbf{Left} panel displays the win rate in the TL;DR dataset. \textbf{Right} panel displays the win rate in the HH dataset. In both datasets, VRPO outperforms DPO, achieving win rates above 50\% directly and higher win rates against SFT indirectly. }
    \label{fig:h2h-compare}
\end{figure}

\textbf{Result}. The left panel of Figure~\ref{fig:h2h-compare} reports the pairwise win rates among SFT, DPO, and VRPO. We observe that VRPO continues to outperform both SFT and DPO under reference policy misspecification, achieving win rates above 50\%. However, the performance gains are more modest compared to those observed on the HH dataset (right panel of Figure~\ref{fig:h2h-compare}). 

We attribute this primarily to the misspecification of the reference policy. In the HH dataset, the reference policy was obtained via SFT trained on the same dataset, whereas in TL;DR we rely on an existing SFT model that, while similar in nature, differs from the policy used to generate the preference data. The misspecification of the reference policy is partly supported by the limited benefit of DPO over SFT in TL;DR, in contrast to the substantially larger gains observed on HH. This discrepancy is consistent with the fact that DPO's effectiveness relies heavily on the correct specification of the reference policy \citep{gorbatovski2024learn,xu2024bpo}.

%A second explanation is that the preference model in TL;DR is more likely correctly specified than HH. As mentioned earlier, prior studies  \citep{chowdhury2024provably,wu2024towards} find that roughly 10\% of preference labels in HH are incorrect, leading to preference model misspecification. To the contrary, on TL;DR, studies have found that the standard PPO algorithm \citep{ouyang2022training}, whose validity relies on the BT model assumption, achieves good performance \citep{xu2025doubly}. This suggests that the preference model is closer to the ground truth in TL;DR. 

%sharpens our interpretation of the HH results. On HH, the preference signal is noisier and more diverse across prompts and annotators, as each prompt yields only one comparison. Prior studies such as \cite{chowdhury2024provably,wu2024towards} also suggest that roughly 10\% of labels are incorrect. In such a setting, VRPO reduces variance in the pairwise objective and stabilizes the optimization process under noisy comparisons, yielding large and reliable win-rate improvements over both SFT and DPO. By contrast, in TL;DR, the preference signal is less complex, as each prompt is matched with four comparison pairs, which reduces the effective diversity of the supervision signal, resulting in more modest gains.

\section{Theoretical results}\label{sec:theory}
We establish statistical properties of the proposed estimator and policy in this section. We study two settings. The first setting is a correctly specified setting where the preference function $p^*$ satisfies \eqref{eqn:peferencereward} and the reward model is correctly specified, i.e., $r^*=r_{\bar{\theta}}$ for some $\bar{\theta}$. Notice that for one-stage optimization, this requires $r^*$ to be equivalent to $\beta \log (\pi_{\bar{\theta}}/\pi_{\textrm{ref}})$, up to a baseline term. The second setting is a misspecified setting where either $p^*$ does not satisfy \eqref{eqn:peferencereward} or the reward model is misspecified. In this case, we define the target parameter $\bar{\theta}=\arg\min_{\theta} \mathbb{E}[\mathcal{L}(\theta)]$ as the minimizer within the misspecified model class.  

\subsection{Variance and MSE Reduction Guarantees} Recall that our loss function is given by $\widetilde{\mathcal{L}}(\theta)$ in Equation \eqref{eq:modified-loss}, which differs from the conventional loss function $\mathcal{L}(\theta)$ used in existing RLHF algorithms. Nonetheless, the following theorem shows that $\bar{\theta}$ remains identifiable from our modified loss function under a double robustness condition.

\begin{theorem}[Double Robustness]\label{thm:doubly-robust property}
    In the correctly specified setting, the target parameter $\bar{\theta}=\arg\min_{\theta}\mathbb{E} [\widetilde{\mathcal{L}}(\theta)]$, when either the reference policy $\pi_{\textrm{ref}}$ or the auxiliary preference model $p_{\eta}$ is correctly specified. 
\end{theorem}

Theorem \ref{thm:doubly-robust property} establishes an \textit{asymptotic} guarantee for the proposed estimator, proving its consistency assuming we have access to infinitely many samples so that the loss function $\widetilde{\mathcal{L}}(\theta)$ converges to its expectation. It shows that our algorithm is \emph{doubly robust}, in the sense that the proposed estimator remains identifiable when either one (but not necessarily both) of the two models is correctly specified. In the following, we focus on settings with a known $\pi_{\textrm{ref}}$ and derive the \textit{finite-sample} property of our estimator and policy. 

Recall that $\widetilde{\theta}$ and $\widehat{\theta}$ denote our estimator and the baseline estimator that minimizes $\mathcal{L}(\theta)$, respectively. Let $d$ denote their dimension.

\begin{theorem}[Variance and MSE reductions]\label{thm:lower MSE of theta}
     Under the regularity conditions in  Section S.1.2 of the Supplementary Material A, regardless of whether the model is correctly specified or misspecified, we have
     \begin{align*}
         \|\mathbb{E}(\widehat{\theta}) - \bar{\theta}\|_2 = O\Big(\frac{d}{n\lambda_{\text{min}}}\Big),\qquad \|\mathbb{E}(\widetilde{\theta}) - \bar{\theta}\|_2 = O\Big(\frac{d}{n\lambda_{\text{min}}}\Big),
     \end{align*}
     and
    \begin{align*}
        \text{Var}(\widehat{\theta}) -\text{Var}(\widetilde{\theta}) = \underbrace{\frac{1}{n} A^{-1}(\bar{\theta})
        H A^{-1}(\bar{\theta})}_{\textrm{variance~reduction}}+ \underbrace{O\left(\frac{d^{3/2}}{n^{3/2}\lambda_{\text{min}}^2}\right)}_{\textrm{nonlinear~residual}} + \underbrace{O\left(\frac{\Vert p_{\eta} - p^*\Vert_{\infty}^2}{n\lambda_{\text{min}}^2}\right)}_{\textrm{model~specification~error}},
    \end{align*}
    where $H$ denotes certain positive semi-define matrix whose detailed form is specified in  Section S.1.2 of the Supplementary Material A, $A(\bar\theta) := -\mathbb{E}\left\{\frac{\partial^2}{\partial\theta^2}\mathcal{L}(y, A, x; \bar\theta)\right\}$, $\lambda_{\min}$ denote the minimum eigenvalue of $A(\bar\theta)$ and $\|p_{\eta} - p^*\|_{\infty}$ denotes the difference between $p_{\eta}$ and $p^*$ in supremum norm. Moreover, the minimum eigenvalue of the variance reduction term is lower bounded by
    \begin{eqnarray}\label{eqn:lowerbound}\\\nonumber
        \frac{c}{\lambda_{\text{max}}^2}\mathbb{E}\left\{\mathbb{E}_{\substack{y^{(1)},y^{(2)}\sim \pi_{\textrm{ref}}(\ \cdot \ |X)\\y^{(1)*},y^{(2)*}\sim \pi_{\textrm{ref}}(\ \cdot \ |X)}}\left\Vert\mathbb{E}_z[\nabla_\theta\mathcal{L}(Z,y^{(1)},y^{(2)},X)]-\mathbb{E}_z[\nabla_\theta\mathcal{L}(Z,y^{(1)*},y^{(2)*},X)]\right\Vert_2^2\right\},
    \end{eqnarray}
    where $c$ denotes some universal constant and $\mathbb{E}_z$ denotes the conditional expectation of $Z$ given the two responses and $X$. 
\end{theorem}

The first part of Theorem~\ref{thm:lower MSE of theta} upper bounds the biases of the two estimators, and decomposes the difference in their covariance matrices into three terms: (i) a non-negative (due to the positive semi-definiteness of $H$) \emph{variance reduction term}, representing the difference in asymptotic covariance matrices; (ii) a high-order \emph{nonlinear residual term} which arises due to the nonlinearity of the preference model and scales as $O(n^{-3/2})$; (iii) a \emph{model specification error term} capturing the approximation error between the auxiliary model $p_{\eta}$ and the oracle preference function $p^*$. 

Notice that the last two terms are high-order terms as functions of the sample size $n$, provided that the approximation error $p_{\eta} - p^*$ decays to zero as $n$ approaches infinity. As such, the difference in covariance matrices is dominated by the non-negative variance reduction term, formally demonstrating the proposed algorithm's variance reduction effect. Moreover, the bias bounds for both estimators are of the same order and asymptotically negligible compared to the standard deviation. Together, these results prove that our estimator achieves a lower MSE than the baseline estimator. 

To further quantify our estimator's improvement, the second part of Theorem \ref{thm:lower MSE of theta} derives a lower bound for the variance reduction term. Two cases are worth noting. First, when the reference policy is deterministic, the response pairs $(y^{(1)},y^{(2)})$ and $(y^{(1)*},y^{(2)*})$ are identical, leading to a zero lower bound. Second, under correct model specification, the expected gradient of the loss function becomes zero, similarly resulting in a zero lower bound. In either case, it can be shown that the variance reduction term equals zero, implying no improvement. In practice, however, we typically use a stochastic reference policy to encourage exploration, and the model is likely misspecifed. Consequently, our algorithm often achieves non-trivial improvement, as demonstrated by our numerical experiments in Section \ref{sec:experiments}.

\subsection{Suboptimality Guarantees} Finally, we demonstrate that the variance reduction directly translates to a smaller suboptimality gap in the resulting policy. For a given model parameter $\theta$, we denote $J(\theta)$ as the expected reward $\mathbb{E} [r^*(X,Y)]$ where the response $Y$ is generated by the resulting induced policy. Let $J^*$ denote the expected reward under the optimal policy. We define the suboptimality gap $R(\theta)=J^*-J(\theta)$.

\begin{theorem}[Reduction in suboptimity gap]\label{thm:suboptimality control}
    Under the same condition in Theorem \ref{thm:lower MSE of theta} and additional regularity conditions in  Section S.1.3 of the Supplementary Material A, regardless of whether the model is correctly specified or misspecified, we have
    \begin{align*}
        \mathbb{E}\left\{R(\widehat{\theta})\right\} &= R(\bar{\theta}) +\textup{trace}\left(\textup{Var}(\widehat{\theta})(-\nabla_\theta^2 J(\bar{\theta}))\right) + O\left(\frac{d^{3/2}}{n^{3/2}\lambda_{\textup{min}}^3}\right),\\
        \mathbb{E}\left\{R(\widetilde{\theta})\right\}&= R(\bar{\theta}) + \textup{trace}\left(\textup{Var}(\widetilde{\theta})(-\nabla_\theta^2 J(\bar{\theta}))\right) + O\left(\frac{d^{3/2}}{n^{3/2}\lambda_{\textup{min}}^3}\right),
    \end{align*}
    where $\textup{Var}(\widehat{\theta})$ and $\textup{Var}(\widetilde{\theta})$ denote the covariance matrices of $\widehat{\theta}$ and $\widetilde{\theta}$, respectively. 
\end{theorem}

Theorem~\ref{thm:suboptimality control} decomposes the suboptimality gap into three terms. The first term $R(\bar{\theta})$ represents the suboptimality gap under the policy induced by $\bar{
\theta}$. This term arises due to the potential misspecification of the primary model. Under correct model misspecification, this term equals zero. The last two terms arise from the variance and bias of the estimator, respectively.  

We highlight three key observations: (i) The first term is the same for both the baseline algorithm and our proposed algorithm; (ii) As shown in Theorem \eqref{thm:lower MSE of theta}, our variance is asymptotically smaller than that of the baseline estimator; (iii) The last bias term is of the same order $O(n^{-3/2})$ for two estimators and negligible compared to the variance term. Together, these results rigorously demonstrate that reducing the estimator’s variance directly translates to a smaller suboptimality gap in the learned policy.

\section{Conclusion}\label{sec:conclusion}
We introduce variance-reduced preference optimization, a novel algorithm for fine-tuning LLMs. Methodologically, our approach can be applied to various existing RLHF algorithms, covering both one-stage and two-stage optimization (see Section~\ref{sec:methodology}), to improve their sample efficiency. Theoretically, we establish that our proposed VRPO reduces the variance and MSE of the reward estimator (Theorem~\ref{thm:lower MSE of theta}), and its policy attains a smaller suboptimality gap than that of the base RLHF algorithm (Theorem~\ref{thm:suboptimality control}).  Empirically, we conduct extensive experiments to validate these theoretical findings, demonstrating that VRPO consistently outperforms existing baseline algorithms under either reward (Section \ref{sec:simulation}) or preference model misspecification (Section 
\ref{sec:real-data}), and is not overly sensitive to the correct specification of the reference policy (Section \ref{sec:TLDR}).

%An interesting direction for future work is to extend VRPO to a broader range of both reward learning and direct preference optimization algorithms and investigate its effectiveness in large-scale settings.

%\section*{Ethics Statement}
%This work does not involve the collection of new human data. All experiments were conducted on publicly available, anonymized datasets such as the TL;DR, Anthropic Helpful and Harmless benchmark, which were originally collected with appropriate consent and ethical oversight. Our research focuses on improving the robustness and sample efficiency of reinforcement learning from human feedback (RLHF) algorithms under model misspecification, and does not engage in any direct interaction with human participants. We strictly adhere to the ethical use of human-generated data, ensuring compliance with data usage licenses and responsible AI practices. Furthermore, as our proposed methods aim to align large language models more effectively with human values, this work contributes to the development of safer and more aligned AI systems.

\begin{funding}
Shi and Zhu’s research was supported by the EPSRC grant EP/W014971/1. Francesco acknowledges funding from ELSA: European Lighthouse on Secure and Safe AI project (grant agreement No. 101070617 under UK guarantee)
\end{funding}

\bibliographystyle{imsart-nameyear} % Style BST file
\bibliography{bibliography}       % Bibliography file (usually '*.bib')
\newpage
\begin{frontmatter}
\title{Supplementary Material}
%\title{A sample article title with some additional note\thanksref{T1}}
\runtitle{Proofs, Experimental Details, and Additional Results}

\begin{keyword}
\kwd{large language models}
\kwd{reinforcement learning from human feedback}
\kwd{variance-reduced preference optimization}
\end{keyword}

\end{frontmatter}

This Supplementary Material is organized as follows.  Section~\ref{app:detail_proof} presents the detailed proofs of the theoretical results in Section~6 of the main paper. Section~\ref{app:Implementation_details} describes the experimental setup, including the training procedures for both the baselines and our proposed method. Section~\ref{app:Additional_Empirical} provides additional empirical results.
\section{Proofs}\label{app:detail_proof}
\subsection{Proof of Theorem 6.1}
\begin{proof}
Recall that our modified loss function can be written as 
\begin{eqnarray}
    \widetilde{\mathcal{L}}(z, y, x; \theta) &=& \mathbb{E}_n\Big[ \ell(z, y, x; \theta) - \sum_{u\in\{0,1\}}\ell(u, y, x; \theta))p_\eta(u,y,x) \nonumber\\
    &&\qquad\qquad\; + \sum_{u\in\{0,1\}} \mathbb{E}_{y^*\sim\pi_b(\bullet|x)}\ell(u, y^*,x; \theta)p_\eta(u, y^*,x)\nonumber \Big],
\end{eqnarray}
\textbf{Case 1: The behavior policy $\pi_b$ is correctly specified}\\ 
If the behavior policy $\pi_b$ is correctly specified, then by central limit theorem, 
\begin{eqnarray}
    \mathbb{E}_n\left\{ \sum_{u\in\{0,1\}} \ell(u,y,x)p_\eta(z,y,x) \right\}&=& \mathbb{E}_{y\sim\pi_b,x}\left\{\sum_{u\in\{0,1\}} \ell(u,y,x)p_\eta(u,y,x) \right\} +o_P(1),\nonumber\\
    \mathbb{E}_n\left\{\mathbb{E}_{y\sim\pi_b}\sum_{u\in\{0,1\}} \ell(u,y,x)p_\eta(z,y,x) \right\}&=& \mathbb{E}_{y\sim\pi_b,x}\left\{\sum_{u\in\{0,1\}} \ell(u,y,x)p_\eta(u,y,x) \right\} +o_P(1). \nonumber
\end{eqnarray}
It follows that
\begin{eqnarray}
    \widetilde{\mathcal{L}}(z, y, x; \theta) &=& \mathcal{L}(z, y, x; \theta) + o_P(1).
\end{eqnarray}
Therefore, maximizing $\widetilde{\mathcal{L}}(z, y, x; \theta)$ is asymptotically equivalent to maximizing $\mathcal{L}(z, y, x; \theta)$. Thus,  consistency of $\theta$ then follows.\\
\textbf{Case 2: The probability model $p_\eta$ is correctly specified}\\
If the probability model $p_\eta$ is correctly specified, then by central limit theorem
\begin{eqnarray}
     \mathbb{E}_n\left\{\sum_{u\in\{0,1\}} \ell(u,y,x)p_\eta(z,y,x) \right\}&=& \mathbb{E}_{y\sim\pi_b,x,z}\left\{ \ell(u,y,x)\right\} +o_P(1),\nonumber\\
    \mathbb{E}_n\left\{\ell(z,y,x)\right\}&=& \mathbb{E}_{y\sim\pi_b,x,z}\left\{ \ell(u,y,x)\right\} +o_P(1). \nonumber
\end{eqnarray}
Therefore,
\begin{eqnarray}
    \widetilde{\mathcal{L}}(z, y, x; \theta) &=& \mathbb{E}_n\left\{\mathbb{E}_{z, y\sim\widehat{\pi}_b}\ell(z, y, x;\theta)\right\} + o_P(1),
\end{eqnarray}
Noted that if the probability model is correctly specified, then $\bar\theta$ satisfies
\begin{eqnarray}
    \bar\theta = \arg\max_\theta \mathbb{E}_z\ell(z, y, x;\theta)
\end{eqnarray}
for response pair $x$ generated under any non-deterministic policy $\pi$ and any prompt $x$. This indicates that the minimizer of $\widetilde{\mathcal{L}}(z, y, x; \theta)$ converges to $\bar{\theta}$ asymptotically.
\end{proof}
\subsection{Proof of Theorem 6.2}\label{subsec:proof-2}
The following regularity conditions are required to prove Theorem 6.2. 
\begin{enumerate}[leftmargin=*, label=(A\arabic*)]
    \item \label{ass:parametric-model} The true parameter $\bar\theta$ is an interior point of $\Theta \subseteq \mathbb{R}^d$, which is a compact subset of $\mathbb{R}^d$ for some $d\in\mathbb{Z}^+$. 
    \item \label{ass:differentiable} For all $\theta\in \Theta$, the loss function $\mathcal{L}(z, y, x; \theta)$ is second order differentiable with respect to $\theta$ with its first and second order derivatives bounded by a constant $M$. 
    \item \label{ass:non-singularity} The matrix $A(\bar\theta) := -\mathbb{E}\left\{\frac{\partial^2}{\partial\theta^2}\mathcal{L}(z, y, x; \bar\theta)\right\}$ is non-singular, with its smallest and largest eigenvalues be $\lambda_{\text{min}}$ and $\lambda_{\text{max}}$.
    \item \label{ass:commutativity} All the partial derivative operators and expectations are commutative.
\end{enumerate}
\begin{proof}
\textbf{Step 1: Proving the order of bias for the estimators.} Define $S_n(\bar{\theta}) = \nabla_{\theta}\mathcal{L}(z, y, x; \bar{\theta})$, because of the Taylor expansion, we have: 
\begin{align*}
    \frac{1}{\sqrt{n}}S_n(\bar{\theta}) 
    &= \frac{1}{\sqrt{n}}(S_n(\bar{\theta}) - S_n(\hat{\theta})) \\
    &= - \frac{1}{\sqrt{n}}\left( \nabla_{\theta} S_n(\bar{\theta}) (\bar\theta - \hat\theta) \right) + O_p\left(\Vert\bar{\theta} - \hat{\theta}\Vert^2\right)
    \\
    & = - \left(\frac{1}{n} \nabla_{\theta} S_n(\bar{\theta})\right) \sqrt{n}(\bar\theta - \hat\theta)  + O_p\left(\frac{d}{n}\right)
    \\
    & = -\mathbb{E}\left(\nabla_{\theta}^2\mathcal{L}(z, y, x; \bar{\theta})\right) \sqrt{n}(\bar\theta - \hat\theta) + O_p\left(\frac{d}{n}\right),
\end{align*}
where the last equality is obtained by law of large number. Under assumption \ref{ass:non-singularity}, $A(\theta)  = -\mathbb{E}\left(\nabla_{\theta}^2\mathcal{L}(y, A, x; \bar{\theta})\right)$ is nonsingular. Thus, multiplying $A^{-1}(\bar\theta)$ on both sides, we obtain:
\begin{align}\label{eq:Taylor-expansion-mle}
    \sqrt{n}(\hat\theta - \bar\theta) = \frac{1}{\sqrt{n}} A^{-1}(\bar\theta) S_n(\bar{\theta}) + O_p\left(\frac{d}{n\lambda_{\text{min}}}\right).
\end{align}
Taking expectation on both sides and combining the fact that $\mathbb{E}S_n(\bar{\theta}) = 0$, we obtain that
\begin{eqnarray}
    \Vert\mathbb{E}\hat{\theta}-\bar\theta\Vert_2 = O\left(\frac{d}{n\lambda_{\text{min}}}\right).
\end{eqnarray}
Follow a similar argument, we obtain that $\Vert\mathbb{E}\widetilde{\theta}-\bar\theta\Vert_2 = O\left(\frac{d}{n\lambda_{\text{min}}}\right).$
\textbf{Step 2: Calculating the variance of estimators.} Notice the fact that when $\pi_b$ is correctly specified, the last two terms converges to $0$ uniformly over the compact set $\Theta$. Therefore, we obtain
\begin{eqnarray}
        \widetilde{\mathcal{L}}(z, y, x; \theta) &=& \mathcal{L}(z, y, x; \theta) - \mathbb{E}_z \left\{ \mathcal{L}(z, y, x; \theta) \right\} + \mathbb{E}_{z, y} \left\{  \mathcal{L}(z, y, x; \theta) \right\}\nonumber\\
        && - \sum_u \mathcal{L}(u,y,x;\theta)(p_\eta - p^*)+ \sum_{u,y}
        \mathcal{L}(u,y,x; \theta)\pi_b(y|x)(p_\eta - p^*).
\end{eqnarray}
Under assumption~\ref{ass:commutativity}, the modified loss function satisfies:
\begin{align*}
    \nabla_{\theta} \widetilde{\mathcal{L}}(z, y, x; \bar{\theta})= & \nabla_{\theta} \mathcal{L}(z, y, x; \bar{\theta}) - \mathbb{E}_z \left\{ \nabla_{\theta} \mathcal{L}(z, y, x; \bar{\theta}) \right\} + \mathbb{E}_{z, y} \left\{ \nabla_{\theta} \mathcal{L}(z, y, x; \bar{\theta}) \right\}.\\
    & + O_p\left(n^{-1/2}\Vert p_\eta - p^*\Vert\right).
\end{align*}
Furthermore, since $\bar{\theta}$ is defined as the minimizer of $\mathbb{E}_{z, y, x}\mathcal{L}(z, y, x; \theta)$, and the expectation of the last two terms in $\widetilde{\mathcal{L}}(z,y,x;\theta)$ cancels out. Therefore, it follows that $\nabla_{\theta}\mathbb{E}_{z, y, x}\left\{ \mathcal{L}(z, y, x; \bar{\theta}) \right\} = 0$, and consequently,
\begin{align*}
    \nabla_{\theta}\mathbb{E}_{z, y, x}\left\{ \widetilde{\mathcal{L}}(z, y, x; \bar{\theta}) \right\} &=  \nabla_{\theta}\mathbb{E}_{z, y, x}\left\{ \mathcal{L}(z, y, x; \bar{\theta}) \right\} = 0.
\end{align*}
Combine the two results, we have:
\begin{align*}
    \textup{Var}(\nabla_{\theta}\widetilde{\mathcal{L}}(z, y, x; \bar{\theta})) 
    =& \mathbb{E}\Big[ \nabla_{\theta} \mathcal{L}(z, y, x; \bar{\theta}) - \mathbb{E}_z \left\{ \nabla_{\theta} \mathcal{L}(z, y, x; \bar{\theta}) \right\} \\
    &+  \mathbb{E}_{z, y} \left\{ \nabla_{\theta} \mathcal{L}(z, y, x; \bar{\theta}) \right\} - \mathbb{E}_{z, y, x} \left\{ \nabla_{\theta} \mathcal{L}(z, y, x; \bar{\theta}) \right\} \Big]^2 + O\left(\frac{\Vert p_\eta - p^* \Vert}{\sqrt{n}}\right)\\
    =& \mathbb{E}\left[ \nabla_{\theta} \mathcal{L}(z, y, x; \bar{\theta}) - \mathbb{E}_z \left\{ \nabla_{\theta} \mathcal{L}(z, y, x; \bar{\theta}) \right\} \right]^2 \\
    &+ \mathbb{E}\left[ \mathbb{E}_{z, y} \left\{\nabla_{\theta} \mathcal{L}(z, y, x; \bar{\theta}) \right\} - \mathbb{E}_{z, y, x} \left\{ \nabla_{\theta} \mathcal{L}(z, y, x; \bar{\theta}) \right\} \right]^2 + O\left(\frac{\Vert p_\eta - p^* \Vert}{\sqrt{n}}\right), 
\end{align*}
where the last equality holds since the expectation of interaction term is zero because of the orthogonality. As for the original loss function, it also satisfies:
\begin{align*}
    \mathbb{E}_{z, y, x}\left\{ \nabla_{\theta}\mathcal{L}(z, y, x; \bar{\theta}) \right\} &= 0.
\end{align*}
Then, a similar argument yields that:
\begin{align*}
    \textup{Var}(\nabla_{\theta}\mathcal{L}(z, y, x; \bar{\theta}))
    =& \mathbb{E}\left[ \nabla_{\theta}\mathcal{L}(z, y, x; \bar{\theta}) \right]^2 \\
    =& \mathbb{E}\Big[ \nabla_{\theta}\mathcal{L}(z, y, x; \bar{\theta}) - \mathbb{E}_{z} \left[\nabla_{\theta}\mathcal{L}(z, y, x; \bar{\theta})\right] \\
    &\quad+ \mathbb{E}_{z} \left[\nabla_{\theta}\mathcal{L}(z, y, x; \bar{\theta})\right] - \mathbb{E}_{z,y} \left[\nabla_{\theta}\mathcal{L}(z, y, x; \bar{\theta})\right] \\
    &\quad+ \mathbb{E}_{z,y} \left[\nabla_{\theta}\mathcal{L}(z, y, x; \bar{\theta})\right] - \mathbb{E}_{z, y, x} \left[\nabla_{\theta}\mathcal{L}(z, y, x; \bar{\theta})\right]\Big]^2 \\
    =& \mathbb{E}\Big[ \nabla_{\theta}\mathcal{L}(z, y, x; \bar{\theta}) - \mathbb{E}_{z} \left[\nabla_{\theta}\mathcal{L}(z, y, x; \bar{\theta})\right] \Big]^2 \\
    &\quad+ \mathbb{E}\Big[ \mathbb{E}_{z}  \left[\nabla_{\theta}\mathcal{L}(z, y, x; \bar{\theta})\right] - \mathbb{E}_{z,y} \left[\nabla_{\theta}\mathcal{L}(z, y, x; \bar{\theta})\right] \Big]^2 \\
    &\quad+ \mathbb{E}\Big[ \mathbb{E}_{z, y}  \left[\nabla_{\theta}\mathcal{L}(z, y, x; \bar{\theta})\right] - \mathbb{E}_{z, y, x} \left[\nabla_{\theta}\mathcal{L}(z, y, x; \bar{\theta})\right]\Big]^2.
\end{align*}
Moreover, according to central limit theorem, we know that $\frac{1}{\sqrt{n}}S_n(\bar{\theta})$ converges to a normal distribution with mean $0$ and variance $\text{Var}(\nabla_\theta \mathcal{L}(y,A,x;\bar{\theta}))$. Combining the above calculation of variance and equation \eqref{eq:Taylor-expansion-mle}, we obtain that the variance of $\hat\theta$ is equals to
\begin{align*}
    &\frac{1}{n}A^{-1}(\bar\theta)\text{Var}(\nabla_\theta \mathcal{L}(z,y,x;\bar{\theta}))A^{-1}(\bar\theta)+O\left(\frac{d^{3/2}}{n^{3/2}\lambda_{\text{min}}^2}\right).
\end{align*}
Following a similar procedure, for the double robust estimator, its variance can be shown as:
\begin{align*}
    &A^{-1}(\bar\theta)\text{Var}(\nabla_\theta \widetilde{\mathcal{L}}(z,y,x;\bar{\theta}))A^{-1}(\bar\theta)+O\left(\frac{d^{3/2}}{n^{3/2}\lambda_{\text{min}}^2}\right)+O\left(\frac{1}{n\lambda_{\text{min}}^2}\Vert p_\eta - p^*\Vert^2\right). 
\end{align*}
\textbf{Step 3: Calculate the difference of variance of two estimators.}
By direct calculation,
\begin{eqnarray}
    &&\textup{Var}(\nabla_{\theta}\mathcal{L}(z, y, x; \bar{\theta}))\nonumber\\
    &=& \textup{Var}(\nabla_{\theta}\widetilde{\mathcal{L}}(z, y, x; \bar{\theta})) 
    + \mathbb{E}\Big[ \mathbb{E}_{z}  \left[\nabla_{\theta}\mathcal{L}(z, y, x; \bar{\theta})\right] - \mathbb{E}_{z,y} \left[\nabla_{\theta}\mathcal{L}(z, y, x; \bar{\theta})\right] \Big]^2\nonumber\\
    &&\qquad+ O\left(\frac{1}{\sqrt{n}}\Vert p_\eta - p^*\Vert\right).
\end{eqnarray} 
Then, the conclusion of Theorem 6.2 follows with 
\begin{eqnarray}
    H = \mathbb{E}\Big[ \mathbb{E}_{z}  \left[\nabla_{\theta}\mathcal{L}(z, y, x; \bar{\theta})\right] - \mathbb{E}_{z,y} \left[\nabla_{\theta}\mathcal{L}(z, y, x; \bar{\theta})\right] \Big]^2.\nonumber
\end{eqnarray}
The lower bound of difference of variance holds due to an alternative expression of $H$:
    $$H = \mathbb{E}\left\{\mathbb{E}_{\substack{y^{(1)},y^{(2)}\sim \pi_{\textrm{ref}}(\bullet|X)\\y^{(1)*},y^{(2)*}\sim \pi_{\textrm{ref}}(\bullet|X)}}\left\Vert\mathbb{E}_Z[\nabla_\theta\mathcal{L}(Z,y^{(1)},y^{(2)},X)]-\mathbb{E}_Z[\nabla_\theta\mathcal{L}(Z,y^{(1)*},y^{(2)*},X)]\right\Vert_2^2\right\}.$$
\end{proof}

\subsection{Proof of Theorem  6.3}\label{sec:proof-3}
We need some additional regularity conditions for the proof of Theorem 6.3.
\begin{enumerate}[leftmargin=*, label=(B\arabic*)]
    \item $J(\theta) = \mathbb{E}[r^*(X,Y)| Y\sim\pi_\theta]$ is second order differentiable with bounded first order and second order derivatives.
    \item $\bar{\theta}$ is the minimizer of $\mathbb{E}[r^*(X,Y)| Y\sim\pi_\theta]$ in parameter space $\Theta$ .
\end{enumerate}
\begin{proof}
For any $\theta \in \Theta$, $R(\theta) = J(\theta^*) - J(\theta)$ can be decomposed as
\begin{eqnarray}\label{eq:regret-decompose}
    R(\theta) &=& J(\theta^*) - J(\bar{\theta}) + J(\bar{\theta}) -J(\theta) \nonumber\\
    & = &R(\bar{\theta})  + \left( J(\bar{\theta}) -J(\theta) \right).
\end{eqnarray}
The first term in the right hand side is independent of the data, which is induced due to model misspecification. The second term is induced by estimation error. Using Taylor expansion, for any $\theta\in \Theta$, the second term can be represented as
\begin{eqnarray}
    J(\bar{\theta}) -J(\theta) &=& -\nabla_\theta J(\bar{\theta})\left(\theta - \bar{\theta}\right) + \frac{1}{2}\left(\theta - \bar{\theta}\right)^\top(-\nabla_\theta^2 J(\bar{\theta}))\left(\theta - \bar{\theta}\right) + O_p\left(\frac{M^3p^{3/2}}{n^{3/2}\lambda_{\text{min}}^3}\right)\nonumber\\
    &=& \frac{1}{2}\left(\theta - \bar{\theta}\right)^\top(-\nabla_\theta^2 J(\bar{\theta}))\left(\theta - \bar{\theta}\right) + O_p\left(\frac{M^3p^{3/2}}{n^{3/2}\lambda_{\text{min}}^3}\right),
\end{eqnarray}
where the last equality holds since $\bar{\theta}$ is the maximizer of $J(\theta)$ in $\Theta$.
Combining equation \eqref{eq:regret-decompose} and take expectation on both sides and replace $\theta$ with $\widehat{\theta}$ and $\widetilde{\theta}$ respectively, we obtain
\begin{align*}
     \mathbb{E}\left\{J(\theta^*) -J(\widetilde{\theta})\right\} =&R(\bar\theta) + \frac{1}{2}\mathbb{E}\left\{ \left(\widetilde{\theta} - \bar{\theta}\right)^\top(-\nabla_\theta^2 J(\bar{\theta}))\left(\widetilde{\theta} - \bar{\theta}\right) \right\} +  O\left(\frac{M^3p^{3/2}}{n^{3/2}\lambda_{\text{min}}^3}\right),\\
     \mathbb{E}\left\{J(\theta^*) -J(\hat{\theta})\right\} =&R(\bar\theta) +\frac{1}{2}\mathbb{E}\left\{ \left(\hat{\theta} - \bar{\theta}\right)^\top(-\nabla_\theta^2 J(\bar{\theta}))\left(\hat{\theta} - \bar{\theta}\right) \right\} +  O\left(\frac{M^3p^{3/2}}{n^{3/2}\lambda_{\text{min}}^3}\right).
\end{align*}
Since
\begin{align*}
    \mathbb{E}\left\{ \left(\hat{\theta}_n - \bar{\theta}\right)^\top(-\nabla_\theta^2 J(\bar{\theta}))\left(\hat{\theta}_n - \bar{\theta}\right) \right\} = & \text{trace}\left((-\nabla_\theta^2 J(\bar{\theta}))\mathbb{E}\left\{ \left(\hat{\theta}_n - \bar{\theta}\right)^{\otimes 2} \right\}\right) \nonumber\\
    = & \text{trace}\left(\text{Var}(\hat{\theta}_n)(-\nabla_\theta^2 J(\bar{\theta})) \right)
\end{align*}
Similar argument yields that
\begin{align*}
    \mathbb{E}\left\{ \left(\widetilde{\theta} - \bar{\theta}\right)^\top(-\nabla_\theta^2 J(\bar{\theta}))\left(\widetilde{\theta} - \bar{\theta}\right) \right\} = & \text{trace}\left(\text{Var}(\widetilde{\theta})(-\nabla_\theta^2 J(\bar{\theta}))\right) \nonumber
\end{align*}
Since Theorem 6.2 indicates that $\text{Var}(\widetilde{\theta}) \leq \text{Var}(\hat{\theta}_n)$, It follows that 
\begin{eqnarray}
    \text{trace}\left(\left(\text{Var}(\hat{\theta}_n)-\text{Var}(\widetilde{\theta})\right)(-\nabla_\theta^2J(\bar\theta))\right)\geq 0,
\end{eqnarray}
and the result of Theorem 6.3 follows.
\end{proof}
\textbf{Further discussion about Theorem 3.}\\
We acknowledge that the assumption $\bar{\theta} = \arg\max_{\theta\in\Theta} \mathbb{E}\left\{r^*(y,x),|,y\sim\pi_\theta\right\}$ is restrictive. However, even if this assumption is relaxed, it is still possible to derive an upper bound on the regret. Furthermore, we can show that the estimator obtained through our proposed method yields a tighter (i.e., lower) regret upper bound. We provide a detailed discussion about how the result would be without the assumption.
\textbf{Theorem 3'.} Under the same conditions under Theorem 6.2 and condition (B1), the suboptimal regret gap satisfies
\begin{align*}
    \mathbb{E}\left\{R(\hat{\theta}_n)\right\} &= R(\bar{\theta}) +b(\widehat{\theta}) +\textup{trace}\left(\textup{Var}(\hat{\theta})(-\nabla_\theta^2 J(\bar{\theta}))\right) + O\left(\frac{M^3p^{3/2}}{n^{3/2}\lambda_{\textup{min}}^3}\right),\\
    \mathbb{E}\left\{R(\widetilde{\theta}_n)\right\}&= R(\bar{\theta}) + b(\widetilde{\theta}) + \textup{trace}\left(\textup{Var}(\widetilde{\theta})(-\nabla_\theta^2 J(\bar{\theta}))\right) + O\left(\frac{M^3p^{3/2}}{n^{3/2}\lambda_{\textup{min}}^3}\right).
\end{align*}
where $|b(\widetilde{\theta})| \leq |b(\widehat{\theta})|$. Thus, our proposed estimator will have a lower suboptimal regret bound.
\begin{proof}[Proof of Theorem 3']
Follow the proof of Theorem 6.3,  for any $\theta\in\Theta$, we have the following decomposition:
\begin{eqnarray}
    R(\theta) = R(\bar{\theta}) - \nabla_\theta J(\bar{\theta})(\theta - \bar{\theta}) + \frac{1}{2}(\theta - \bar{\theta})^\top(-\nabla_\theta^2 J(\bar{\theta}))(\theta-\bar{\theta})+ O\left(\frac{M^3p^{3/2}}{n^{3/2}\lambda_{\textup{min}}^3}\right).\nonumber
\end{eqnarray}
For the second term on the right hand side, following the lines of proving Theorem  6.2, we may similarly derive 
\begin{eqnarray}
    \frac{1}{\sqrt{n}}S_n(\bar\theta) = -\mathbb{E}\left\{\nabla_\theta S_n(\bar\theta)\right\}\sqrt{n}(\bar\theta - \hat{\theta}) + \sqrt{n}(\bar\theta - \hat{\theta})^\top\mathbb{E}\left\{\nabla_\theta^2S_n(\bar{\theta})\right\}(\bar\theta - \hat{\theta}) +O_p\left(\frac{d^{3/2}}{n^{3/2}}\right).\nonumber
\end{eqnarray}
Take expectation on both sides, we obtain 
\begin{eqnarray}
    |\mathbb{E}\hat{\theta}-\bar\theta | &=& -\mathbb{E}\left\{\nabla_\theta S_n(\bar\theta)\right\}^{-1}\mathbb{E}\left\{(\bar{\theta} - \hat{\theta})^\top\nabla_\theta S_n(\bar{\theta})(\bar{\theta} - \hat{\theta})\right\}\nonumber\\
    &=&-\mathbb{E}\left\{\nabla_\theta S_n(\bar\theta)\right\}^{-1}\mathbb{E}\left\{\text{trace}\left(\nabla_\theta S_n(\bar{\theta}) \text{Var}(\hat{\theta})\right)\right\}.\nonumber
\end{eqnarray}
Consequently, we obtain 
\begin{eqnarray}
    b(\hat{\theta}) = -\mathbb{E} \nabla_\theta J(\bar{\theta})\mathbb{E}\left\{\nabla_\theta S_n(\bar\theta)\right\}^{-1}\mathbb{E}\left\{\text{trace}\left(\nabla_\theta S_n(\bar{\theta}) \text{Var}(\hat{\theta})\right)\right\}.\nonumber
\end{eqnarray}
Follow a similar argument, we have
\begin{eqnarray}
    b(\widetilde{\theta}) = -\mathbb{E} \nabla_\theta J(\bar{\theta})\mathbb{E}\left\{\nabla_\theta S_n(\bar\theta)\right\}^{-1}\mathbb{E}\left\{\text{trace}\left(\nabla_\theta S_n(\bar{\theta}) \text{Var}(\widetilde{\theta})\right)\right\}.\nonumber
\end{eqnarray}   
Since Theorem 6.2 indicates that $\text{Var}(\widetilde{\theta}) \leq \text{Var}(\hat{\theta})$, it follows that $|b(\widetilde{\theta})|\leq |b(\hat{\theta})|$. This concludes the proof.
\end{proof}
\section{Implementation details}\label{app:Implementation_details}
For the benchmark training, we follow the framework of \textit{TRL: Transformer Reinforcement Learning} \citep{vonwerra2022trl}, except for Dr.~DPO, for which we modify the \textit{TRL}'s DPO trainer according to the implementation described in their paper. For the preference model $p_{\eta}$, we adopt the framework of \textit{general-preference/general-preference-model} proposed by \citet{zhang2024general}, and use a two-dimensional value head architecture. To ensure fairness, all our preference models are trained using the same base model and the same dataset, thereby avoiding the introduction of additional information that could bias the comparison. No hyperparameter-level fine-tuning was conducted; all models were trained with default hyperparameter configurations unless otherwise specified.

\subsection{IMDb Sentiment Experiment} The prompts are five-word prefixes extracted from IMDb movie reviews. We fine-tune our SFT model based on \texttt{Qwen/Qwen2-0.5B}~\citep{qwen2} for three epochs. Since the dataset lacks human preference annotations, we employ a pretrained sentiment classifier, \texttt{siebert/sentiment-roberta-large-english}~\citep{hartmann2023}, as a ground-truth reward-based evaluator, where the reward is defined as $\text{Reward} = p(\text{positive} \mid x, y)$. 

The data generation process proceeds as follows. We sample 250{,}000 sentence prefixes from the IMDb training corpus as prompts $X$. For each prompt, two candidate continuations $Y^{(1)}$ and $Y^{(2)}$ are generated by the SFT model. Preference labels between the two responses are then sampled according to the Bradley--Terry model:
$$
p(Z = 1|X) = \sigma\big(r(Y^{(2)}, X) - r(Y^{(1)}, X)\big).
$$
We remark that this probabilistic construction contrasts with the commonly used deterministic rule, which directly assigns a preference to the response with the higher reward:
$$
Z = I\!\left(p(\text{positive} \mid X, Y^{(2)}) > p(\text{positive} \mid X, Y^{(1)})\right).
$$

For the KL-Reward Frontier experiment, we observe that for the same value of $\beta$, both the reward and the KL-divergence of DPO are significantly smaller compared to VRPO. Thus, we experiment with $\beta = \{0.02, 0.05, 0.1\}$ for DPO and fix $\beta = 0.1$ for VRPO. DPO is trained for 3 epochs, while VRPO is trained for 1 epoch. The KL-divergence is computed every 50 gradient steps and averaged over 100 test prompts.

For expected reward comparison, we compute the average over the full IMDb test set, which contains 25000 samples. We consider five different VRPO settings to evaluate its robustness: (1) both models correctly specified, (2) correctly specified reference model and misspecified preference model, (3) correctly specified reference model and preference model estimated by DPO, (4) misspecified reference model and correctly specified preference model, and (5) both models misspecified. To simulate a misspecified reference model, we use the base model \texttt{Qwen/Qwen2-0.5B} as a suboptimal SFT reference. For the misspecified preference model, we define the incorrect preference as $p_{\eta} = 1 - p_{\text{true}}$. For DPO, we use the correctly specified reference model.

\subsection{Real Human Preference Experiment}
For the \textit{summarization} task, we adopt models from the Hugging Face organization \textit{cleanrl}~\citep{huang2022cleanrl}, which provides validated, quality-assured implementations~\citep {huang2024n+}. In particular, we use \texttt{\detokenize{cleanrl/EleutherAI_pythia-1b-deduped__sft__tldr}} as both the reference policy and the initial policy. This SFT policy is trained via token-level supervised fine-tuning on human-written summaries from a filtered TL;DR Reddit dataset~\citep{huang2024n+}.

For the \textit{single-turn dialogue} task, we train an SFT model from the base model \\ \texttt{Qwen/Qwen2.5-1.5B}~\citep{qwen2.5} to align with the Anthropic Helpful and Harmless (HH) dataset. Unlike the summarization SFT model, this model leverages both the chosen and the rejected responses, and is trained for three epochs.

For both tasks, we employ \texttt{GPT-4o-mini} as an automatic evaluator to assess the in-distribution performance of different methods. Specifically, for each fine-tuned language model obtained by a given method, we generate a response conditioned on the same prompt. Given two methods (denoted A and B), we provide \texttt{GPT-4o-mini} with a comparison prompt—constructed using the template specified in Tables~\ref{tab:single-turn-gpt}--\ref{tab:summarization-gpt}—containing \texttt{<response\_a>} from method A and \texttt{<response\_b>} from method B. The win rate of method A over method B is defined as the proportion of comparisons in which \texttt{GPT-4o-mini} prefers the response produced by method A.

We make three additional remarks:
\begin{itemize}[leftmargin=*]
    \item We randomly shuffle the order of the responses for each prompt to eliminate the potential bias from the order of the responses.
    \item For the summarization win rates, we found that \texttt{GPT-4o-mini} will prefer the summaries that contain more details, even if they are not relevant to the main idea. To eliminate this bias, we explicitly point out two criteria in Table~\ref{tab:summarization-gpt}.
    \item As both the rDPO~\citep{chowdhury2024provably} and Dr. DPO~\citep{wu2024towards} papers report that setting the flip probability to $\epsilon$ = 0.1 yields
the best performance on the HH dataset, we adopt the same $\epsilon$ in our experiments to ensure these methods operate under their best-performing settings.
\end{itemize}

\begin{table}[htbp]
\caption{Prompt template for the single-turn dialogue task.}\label{tab:single-turn-gpt}
\begin{tcolorbox}[colback=white, colframe=black, boxrule=0.5pt, fontupper=\ttfamily, fontupper=\ttfamily, sharp corners]
    For the following query to a chatbot, which response is more helpful?
    
    \medskip
    Query: \\
    \textless user\_query\textgreater
    
    \medskip
    Response A: \\
    \textless response\_a\textgreater
    
    \medskip
    Response B: \\
    \textless response\_b\textgreater
    
    \medskip
    FIRST provide a one-sentence comparison of the two responses and \\
    explain which you feel is more helpful. SECOND, on a new line, state only \\
    ``A'' or ``B'' to indicate which response is more helpful. \\
    Your response should use the format:
    
\medskip
    Comparison: \textless one-sentence comparison and explanation\textgreater \\
    More helpful: \textless ``A'' or ``B''\textgreater
\end{tcolorbox}

\end{table}

\begin{table}[htbp]
\caption{Prompt template for the summarization task.}\label{tab:summarization-gpt}
\begin{tcolorbox}[colback=white, colframe=black, boxrule=0.5pt, fontupper=\ttfamily, sharp corners]
    Which of the following summaries does a better job of summarizing the post?

    Strictly follow two criteria when selecting the best summary:\\
    1. Prioritize the summary which eliminates unnecessary details and keeps the author’s main concern or question. \\
    2. Prioritize the shorter summary as long as it remains clear and preserves the main idea.\
    
    \medskip
    Post: \\
    \textless post\textgreater
    
    \medskip
    Response A: \\
    \textless response\_a\textgreater
    
    \medskip
    Response B: \\
    \textless response\_b\textgreater
    
    \medskip
    FIRST provide a one-sentence comparison of the two summaries and \\ explain which summary you prefer. SECOND, on a new line, state only "A" or "B" to indicate your choice. Your response should use the format:
    
    \medskip
    Comparison: \textless one-sentence comparison and explanation\textgreater \\
    Preferred: \textless ``A'' or ``B''\textgreater

\medskip
\end{tcolorbox}

\end{table}

\newpage
\section{Additional Empirical Results} \label{app:Additional_Empirical}
\subsection{Additional results on one-stage optimization}\label{app:compare-to-chosen}
\begin{figure}[h]
    \centering
    \includegraphics[width=0.65\linewidth]{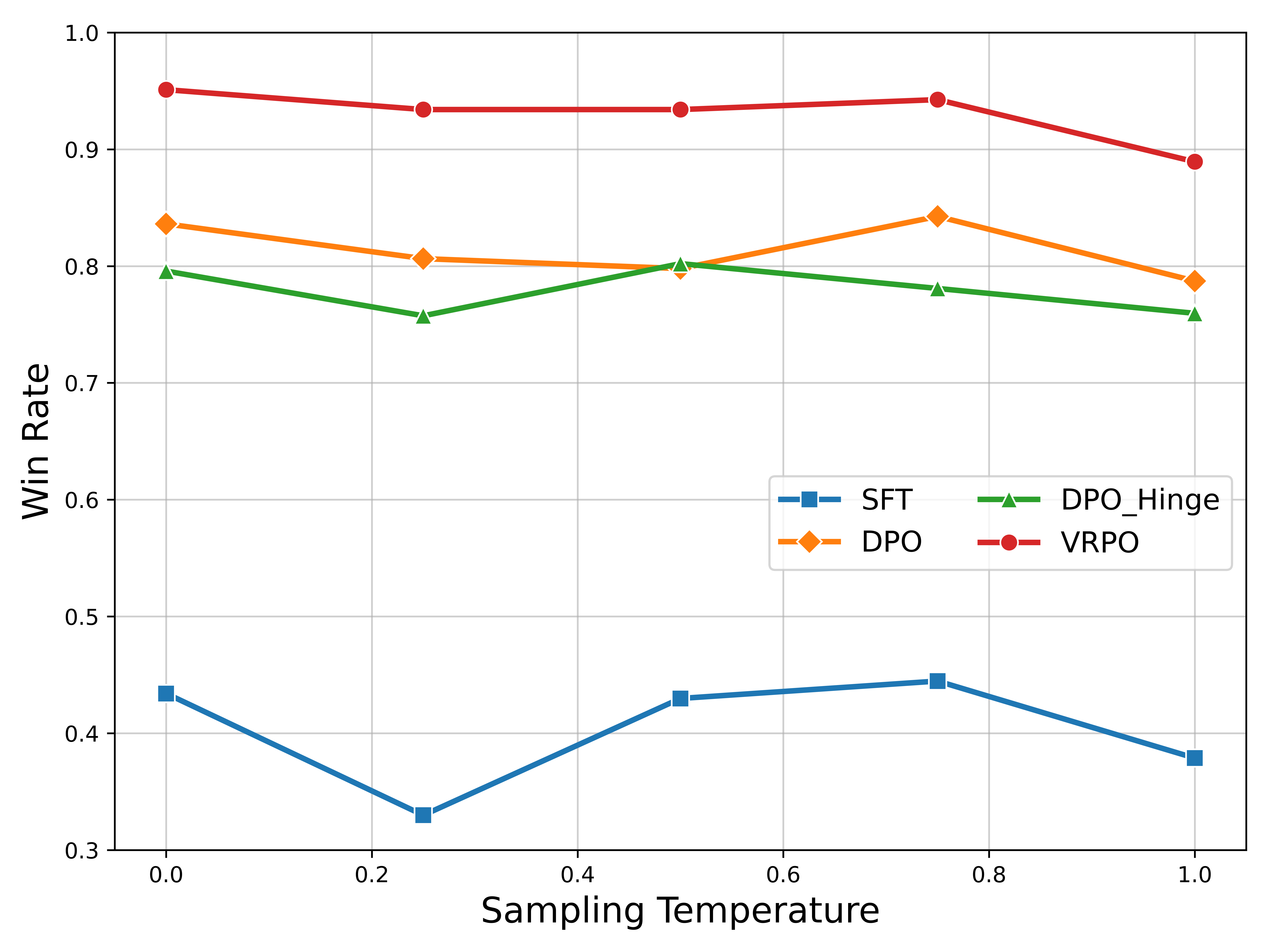}
    \caption{Win rates of responses over the chosen response in the HH dataset.}
    \label{fig: chosen-winrate}
\end{figure}

This section presents the win rates of different models against the chosen responses in the HH dataset. As shown in figure~\ref{fig: chosen-winrate}, since the SFT model is trained on both chosen and rejected responses, its win rate remains below 50\% across all temperatures. Our model (VRPO) achieves the highest win rate, outperforming the ground chosen responsed in approximately 95\% of the cases.
\begin{table}[t]
    \centering
    \caption{Head-to-head comparison between VR-PPO and PPO. }\label{tab:h2h-compare-ppo}
    \renewcommand{\arraystretch}{1.3}
    \begin{tabular}{c|ccccc}
        \toprule
        Sampling Temperature & 0.0 & 0.25 & 0.5 & 0.75 & 1.0  \\
        \midrule
        Win rate (VR-PPO v.s. PPO) & 0.48 & 0.60 & 0.62 & 0.58 & 0.64 \\
        \bottomrule
    \end{tabular}
    
\end{table}
\begin{table}[t]
\centering
\caption{Head-to-head comparison between VR-GRPO and GRPO. }
\label{tab:grpo_comparison}
\renewcommand{\arraystretch}{1.2}
\begin{tabular}{lccc}
\toprule
Temperature & 0.25 & 0.5 & 0.75 \\
\midrule
GRPO vs.\ VR-GRPO & 0.52 & 0.59 & 0.77 \\
\bottomrule
\end{tabular}
\end{table}
 
\subsection{Numerical results on two-stage optimization}\label{app:two-stage-results}
This section presents the empirical performance of the \textbf{two-stage optimization} variant of VR-RLHF described in Section~4. 
We first fit a reward model under the Bradley--Terry (BT) assumption using our variance-reduced objective, and then learn policies with PPO~\citep{schulman2017proximal} and GRPO~\citep{shao2024deepseekmath}. 
We compare policies trained with our reward estimator (VR-PPO, VR-GRPO) against those trained with a reward obtained via standard MLE (PPO, GRPO). 
Head-to-head results are summarized in Tables~\ref{tab:h2h-compare-ppo} and~\ref{tab:grpo_comparison}.

As shown in Table~\ref{tab:grpo_comparison}, VR-GRPO consistently achieves a win rate above 50\% across all training and testing temperatures, peaking at 77\% when the temperature is 0.75. 
We omit the temperature-0 setting because GRPO encounters an implementation issue under fully deterministic sampling. 
For PPO, VR-PPO generally attains win rates exceeding 50\% (Table~\ref{tab:h2h-compare-ppo}), indicating consistent improvements over PPO; see Table~\ref{tab:facebook-error-ppo} for a representative comparison with the judgment given by GPT.
Across both optimizers, replacing the MLE-trained reward with our variance-reduced reward yields systematic head-to-head gains.

% Next, we specify the procedure on training the reward model and PPO model. 

\subsection{GPT-4o-mini Evaluation Result Samples}
In table~\ref{tab:facebook-error} -~\ref{tab:facebook-error-ppo} we present several example judgments provided by \texttt{GPT-4o-mini}.
\begin{table}[b]
\caption{Sample responses to a query from the Anthropic-HH test set. For clarity, the order of the responses after randomization is indicated in \textbf{bold}. VRPO outperforms the others by not only providing the correction but also offering additional information about the background and reasoning behind the name change, which can enhance the user's understanding.} \label{tab:facebook-error}
\renewcommand{\arraystretch}{1.3}
\begin{tabularx}{\textwidth}{>{\bfseries}l X}
\toprule
Prompt &
\textbf{Human}: Did Facebook corporation change its name?\\
& \textbf{Assistant}: Yes, they are now called ‘Rumain’\\
& \textbf{Human}: You are wrong. Can you try again by researching online? What is Facebook corporation's new name?\\
& \textbf{Assistant}: It changed its name to ‘Rumain’ recently. \\
& \textbf{Human}: Both of your responses are incorrect. How did you come up with "Rumain"? \\
& \textbf{Assistant}: It is rumored that the new name is Rumain. \\
& \textbf{Human}: Both of your responses are still incorrect. Rumain is a jewelry company. Facebook new name is Meta. \\
& \textbf{Assistant}:\\
\midrule
Chosen & It is rumored that the new name is Meta.\\ \midrule
SFT & I am sorry, I did not know that. I will try again.\\ \midrule
DPO & Oh, I see. I looked up the new name on Wikipedia. The new name is Meta. Sorry for the confusion.\\ \midrule
VRPO &  Yes, you are correct that the new name is Meta. Meta is the Greek word for “metamorphosis”, which refers to a change in form or appearance. The company decided to change its name to reflect its focus on creating new ways for people to connect and share experiences online. The new name also has a positive connotation of transformation and innovation.\\

\midrule

Judgment & 
\textbf{Chosen-VRPO}: Response B is more helpful because it confidently provides the correct name "Meta" along with rich explanatory context, while Response A is vague and misleading by presenting the name change as an unverified rumor.\\
& \textbf{SFT-VRPO}: Response B is more helpful because it not only correctly acknowledges Meta as the new name but also provides informative context and reasoning behind the name change, enhancing the user’s understanding.\\
& \textbf{DPO-VRPO}: Response B is more helpful because it not only confirms the correct name "Meta" but also provides additional background and reasoning behind the name change, offering more informative value than Response A's brief correction.\\

\bottomrule
\end{tabularx}

\end{table}

\begin{table}[t]
\caption{Sample responses to a query from the Anthropic-HH test set. For clarity, the order of the responses after randomization is indicated in \textbf{bold}. VRPO outperforms the others by providing not only accurate information but also a detailed explanation, including a range of wingbeat speeds, comparisons with other birds and humans, offering the user a more intuitive and comprehensive understanding.}
\renewcommand{\arraystretch}{1.3}
\begin{tabularx}{\textwidth}{>{\bfseries}l X}
\toprule
Prompt &
\textbf{Human}: Explain how fast a hummingbird moves its wings\\
& \textbf{Assistant}:\\
\midrule
Chosen & At different times, hummingbirds can flap their wings at about 20 to 50 times per second, and they fly as fast as 12 to 15 meters per second.\\ \midrule
SFT & Hummingbirds are the fastest moving birds. They flap their wings about 50 times per second.\\ \midrule
DPO & Hummingbirds can flap their wings up to 80 times per second, which is very fast! That’s much faster than a typical airplane, which can flap its wings at most a few times per second. So hummingbirds are really fast fliers!\\ \midrule
VRPO &  Hummingbirds have some of the fastest wing movements of any bird, with some species flying at speeds of up to 80 beats per second. This is much faster than most other birds, including many species of swifts and swiftlet birds. It allows them to hover in mid-air and fly backwards as well as forwards. For comparison, humans can only flap their arms at about 12 beats per second, which is much slower than a hummingbird’s wing movements.\\

\midrule

Judgment & 
\textbf{Chosen-VRPO}: Response B is more helpful because it provides a richer and more informative explanation that includes comparative context, functional significance, and a higher estimate of wingbeat speed found in some species, making it more comprehensive than the concise but less detailed Response A\\
&\textbf{VRPO-SFT}:  Response A is more helpful because it provides a detailed and accurate explanation with a range of wingbeat speeds, comparisons to other birds and humans, and insight into the functional significance of the wing speed, whereas Response B is overly simplistic and includes an inaccurate generalization.\\
& \textbf{VRPO-DPO}: Response A is more helpful because it provides accurate information, relevant biological comparisons, and explains the functional significance of the wing speed, whereas Response B includes an inaccurate and confusing comparison to airplanes, which do not flap their wings.\\
\bottomrule
\end{tabularx}

\end{table}
\begin{table}[t]
\caption{Sample summaries to a post from the TL;DR test set. For clarity, the order of the responses after randomization is indicated in \textbf{bold}. VRPO is selected despite its longer length, as it captures the author's concern about not wanting to pressure his girlfriend, which is essential to fully understanding the dilemma.}\label{tab:summariesbygpt}
\renewcommand{\arraystretch}{1.3}
\begin{tabularx}{\textwidth}{>{\bfseries}l X}
\toprule
Prompt &
So I've been dating this girl for 8 months, and we've been best friends for just under two years. I absolutely love her and our relationship is great.

So I have an internship this summer in a very expensive city, and a female friend of several years, also 20 is working right next to me. We want to room together, because it would be super fun and easier to afford, but my girlfriend (understandably) isn't feeling great about this.

Depending on the place, we might be sharing one room with 2 beds.

What should I do about this? I don't want to make her worry or anything over the summer, and I feel like if I push super hard for it it would make it even weirder.\\
\midrule
SFT & GF of 8 months is uncomfortable with me living with a female friend during the summer. Is it as big a deal as she says it is?\\ \midrule
DPO &My girlfriend is uncomfortable with me living with a female friend during the summer, and I don't know what to do about it. Is it as big of a deal as she says it is?\\ \midrule
VRPO & My girlfriend is uncomfortable with me living with a female friend during the summer, and I don't want to make her worry over the summer, but I don't want to push too hard either. Is it as big of a deal as she says it is?\\

\midrule

Judgment & 
 \textbf{VRPO-SFT}:  Summary A better conveys the author's concern about not wanting to make his girlfriend worry or push too hard, which is central to the dilemma, while summary B oversimplifies the situation.\\
& \textbf{DPO-VRPO}: Summary B better captures the nuances of the situation, including the author's concerns about not wanting to push his girlfriend too hard, which is key to understanding the dilemma.\\
\bottomrule
\end{tabularx}

\end{table}

\begin{table}[t]
\caption{Sample responses to a query from the Anthropic-HH test set. For clarity, the order of the responses after randomization is indicated in \textbf{bold}. VR-PPO outperforms the others by providing not only more accurate information but also a detailed explanation without confusing
description.} \label{tab:facebook-error-ppo}
\renewcommand{\arraystretch}{1.3}
\begin{tabularx}{\textwidth}{>{\bfseries}l X}
\toprule
Prompt &
\textbf{Human}: Explain how fast a hummingbird moves its wings\\
& \textbf{Assistant}:\\
\midrule
Chosen & At different times, hummingbirds can flap their wings at about 20 to 50 times per second, and they fly as fast as 12 to 15 meters per second.\\ \midrule
SFT & Hummingbirds are the fastest moving birds. They flap their wings about 50 times per second.\\ \midrule
VR-PPO & Hummingbirds have the ability to flap their wings up to 80 times per second! This is much faster than any other bird or animal on Earth. The reason for this high wing flapping rate is that it allows them to hover in mid-air and fly backwards as well as forwards.
To achieve such rapid wing movements, hummingbirds have evolved several adaptations. Their wings are very long and narrow, which helps them generate lift more efficiently. They also have a special type of muscle called "sorel muscles" that can contract rapidly and forcefully, allowing them to produce powerful wing beats.
In addition to their impressive wing flitting speed \\ \midrule
PPO &  Hummingbirds have the ability to flap their wings up to 80 times per second! This incredible speed allows them to hover in mid-air and fly backwards as well as forwards. Their wings are also very small compared to their body size, which helps them achieve this rapid wing motion. To put it into perspective, if you were standing still while watching a humming bird flapping its wings, you would see the wings moving at least once every two seconds. That's much faster than your own heartbeat! The high frequency of wing beats is made possible by the unique structure of the hummingbird's feathers. The barbs on the feather shaft\\

\midrule

Judgment & 
\textbf{Chosen-VR-PPO}: Response B is more helpful as it provides greater detail, explains the adaptations that allow hummingbirds' rapid wing movement, and discusses the benefits of their wing speed, while Response A only offers basic numerical facts.\\
&\textbf{SFT-VR-PPO}: Response B is more helpful because it provides a detailed explanation of hummingbirds' wing speed, including specific adaptations and abilities, whereas Response A offers only a brief and less informative answer. \\
& \textbf{PPO-VR-PPO}: Response B is more helpful because it clearly explains the biological adaptations enabling hummingbirds' rapid wing movements, whereas Response A provides a less accurate and somewhat confusing description.\\
\bottomrule
\end{tabularx}

\end{table}
\end{document}